\documentclass{article}

\usepackage[numbers]{natbib}
\usepackage[preprint]{flashvideog}
\usepackage{graphicx}
\usepackage[dvipsnames]{xcolor}         % colors
\definecolor{linkColor}{rgb}{0.18,0.39,0.62}
\usepackage[utf8]{inputenc} % allow utf-8 input
\usepackage[T1]{fontenc}    % use 8-bit T1 fonts
\usepackage[colorlinks=true,linkcolor=linkColor,citecolor=linkColor,filecolor=linkColor,urlcolor=linkColor]{hyperref}       % hyperlinks
\usepackage{url}            % simple URL typesetting
\usepackage{booktabs}       % professional-quality tables
\usepackage{amsfonts}       % blackboard math symbols
\usepackage{nicefrac}       % compact symbols for 1/2, etc.
\usepackage{microtype}      % microtypography

\usepackage{graphicx}
\usepackage{url}            % simple URL typesetting
\usepackage{booktabs}       % professional-quality tables
\usepackage{nicefrac}       % compact symbols for 1/2, etc.
\usepackage{microtype}      % microtypography
\usepackage{wrapfig}
\usepackage{enumitem}
\usepackage{amsmath,amsfonts,bm}
\definecolor{earthyellow}{RGB}{225, 169, 95}

\bibpunct{[}{]}{;}{a}{}{,}

\usepackage{algpseudocode}% http://ctan.org/pkg/algorithmicx
\usepackage{multirow}
\usepackage{amsmath}
\usepackage{capt-of}
\usepackage{makecell}
\usepackage{romannum}
\usepackage{tabularx}
\usepackage{epsfig}
\usepackage{amssymb}
\usepackage{amsfonts}
\usepackage{subcaption} % to use subcaption for subtable
\usepackage{booktabs}
\usepackage{scalerel}
\usepackage{xfrac}
\usepackage{varwidth}
\usepackage[export]{adjustbox}
\usepackage{tikz}
\usetikzlibrary{tikzmark}

\usepackage{stmaryrd}
\usepackage{bbm}
\usepackage{wrapfig}
\usepackage{pifont}

\definecolor{mygreen}{rgb}{0.0, 0.5, 0.0}  % Define your custom green
\definecolor{myblue}{rgb}{0.0, 0.0, 1.0}   % Define your custom blue
\definecolor{deepblue}{rgb}{0,0,0.5}
\definecolor{officeblue}{RGB}{0,102,204}
\definecolor{deepred}{rgb}{0.6,0,0}
\definecolor{deepgreen}{rgb}{0,0.5,0}
\definecolor{mybrickred}{RGB}{182,50,28}

\definecolor{fillcolor}{RGB}{216,217,252}

\newcommand{\supp}{\textit{Supplementary Materials}\xspace}

\usepackage{pifont}% http://ctan.org/pkg/pifont

\usepackage{amssymb}
\newcommand\ours{\textit{FlashVideo}}

\definecolor{ourscolor}{gray}{.9}

\usepackage{amsmath,amsfonts,bm}
\usepackage{pifont}

\def\eqref#1{(\ref{#1})}
\def\1{\bm{1}}

\DeclareMathAlphabet{\mathsfit}{\encodingdefault}{\sfdefault}{m}{sl}
\SetMathAlphabet{\mathsfit}{bold}{\encodingdefault}{\sfdefault}{bx}{n}

\usepackage{graphicx}
\usepackage{url}            % simple URL typesetting
\usepackage{booktabs}       % professional-quality tables
\usepackage{nicefrac}       % compact symbols for 1/2, etc.
\usepackage{microtype}      % microtypography
\usepackage{wrapfig}

\usepackage{enumitem}
\usepackage[ruled,noend,linesnumbered]{algorithm2e}

\usepackage{multirow}
\usepackage{tablefootnote}

\usepackage{color}
\usepackage{colortbl}
\usepackage{xcolor}

\definecolor{blgrey}{rgb}{0.6,0.6,0.6}
\definecolor{bblue}{rgb}{0.855,0.933,0.98}
\definecolor{dblue}{HTML}{5297D6}
\definecolor{gainred}{rgb}{0.1,0.5,0.3}
\definecolor{citecolor}{HTML}{0071BC}
\definecolor{linkcolor}{HTML}{ED1C24}

\usepackage{listings}
\usepackage{color}

\definecolor{dkcyan}{cmyk}{1,0,0,.25}
\definecolor{dkgreen}{rgb}{0,0.6,0}
\definecolor{gray}{rgb}{0.5,0.5,0.5}
\definecolor{mauve}{rgb}{0.58,0,0.82}

\title{\includegraphics[scale=0.55]{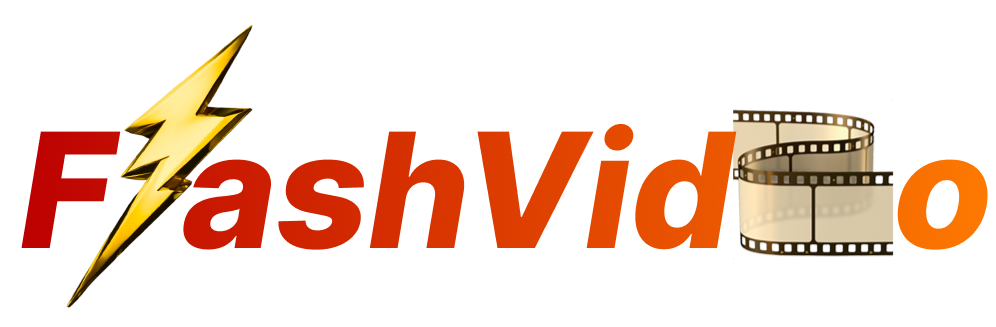} \\ 
\vspace{-3mm}
Flowing Fidelity to Detail for Efficient High-Resolution Video Generation
\vspace{4mm}
}

\author{
  \vspace{-15mm}\\
  \textbf{Shilong Zhang$^{1}$\thanks{:Equal Contribution, $\dag$: project leader }
 \quad   Wenbo Li$^{2*}$
  \quad Shoufa Chen$^{1}$
  \quad Chongjian GE$^{1}$
  } \vspace{2mm} \\
  \textbf{
  Peize Sun$^{1}$
       \quad  Yifu Zhang$^{3}$
  \quad Yi Jiang$^{3\dag}$
 \quad  Zehuan Yuan$^{3}$
  \quad  Bingyue Peng$^{3}$
\quad Ping Luo$^{1}$}\vspace{2mm} \\
  $^1$The University of Hong Kong~~\quad $^2$The Chinese University of Hong Kong ~~\quad  $^3$ByteDance \vspace{2mm} \\
  Code \& Model:~\, \url{https://github.com/FoundationVision/FlashVideo} \\
  \vspace{-5mm} \\
}
\begin{document}
\maketitle

\begin{abstract}
{DiT models have achieved great success in text-to-video generation, leveraging their scalability in model capacity and data scale. High content and motion fidelity aligned with text prompts, however, often require large model parameters and a substantial number of function evaluations (NFEs). Realistic and visually appealing details are typically reflected in high-resolution outputs, further amplifying computational demands-especially for single-stage DiT models. To address these challenges, we propose a novel two-stage framework, FlashVideo, which strategically allocates model capacity and NFEs across stages to balance generation fidelity and quality. In the first stage, prompt fidelity is prioritized through a low-resolution generation process utilizing large parameters and sufficient NFEs to enhance computational efficiency. 
The second stage achieves a nearly straight ODE trajectory between low and high resolutions via flow matching, effectively generating fine details and fixing artifacts with minimal NFEs. To ensure a seamless connection between the two independently trained stages during inference, we carefully design degradation strategies during the second-stage training. Quantitative and visual results demonstrate that FlashVideo achieves state-of-the-art high-resolution video generation with superior computational efficiency. Additionally, the two-stage design enables users to preview the initial output and accordingly adjust the prompt before committing to full-resolution generation, thereby significantly reducing computational costs and wait times as well as enhancing commercial viability. 
}

\end{abstract}

\section{Introduction}

In recent years, text-to-video (T2V) generation has achieved remarkable progress, driven by advances in diffusion probabilistic modeling~\citep{diffusion, ho2020denoising, rectifiedflow, fm}, cutting-edge architectures~\citep{unet, dit}, and the integration of extensive model parameters and large-scale datasets~\citep{lvdm, cogvideo, videocrafter1, videocrafter2, videopoet, opensora, cogvideox, sora}. Among these, DiT-based models~\citep{dit} stand out for their excellent scalability in accommodating larger model capacities and datasets. 

In video DiTs, the key operator is the 3D full attention mechanism across time ($T$), height ($H$), and width ($W$), which effectively models visual relations in scenarios with large object motions and 3D consistency. The computational complexity scales as $\mathcal{O}(T^2 H^2 W^2 \cdot C \cdot N)$, where $C$ represents the feature dimension (linked to model size) and $N$ is the number of denoising steps (function evaluation). State-of-the-art methods~\citep{moviegen, kong2024hunyuanvideo, cogvideox} typically require large model capacities (\textit{e.g.}, 12 billion parameters), high-resolution modeling (\textit{e.g.}, 1080p), and up to 50 denoising steps, for high-quality outputs. 

\begin{figure}[t]
\begin{center}
\includegraphics[width=1\linewidth]{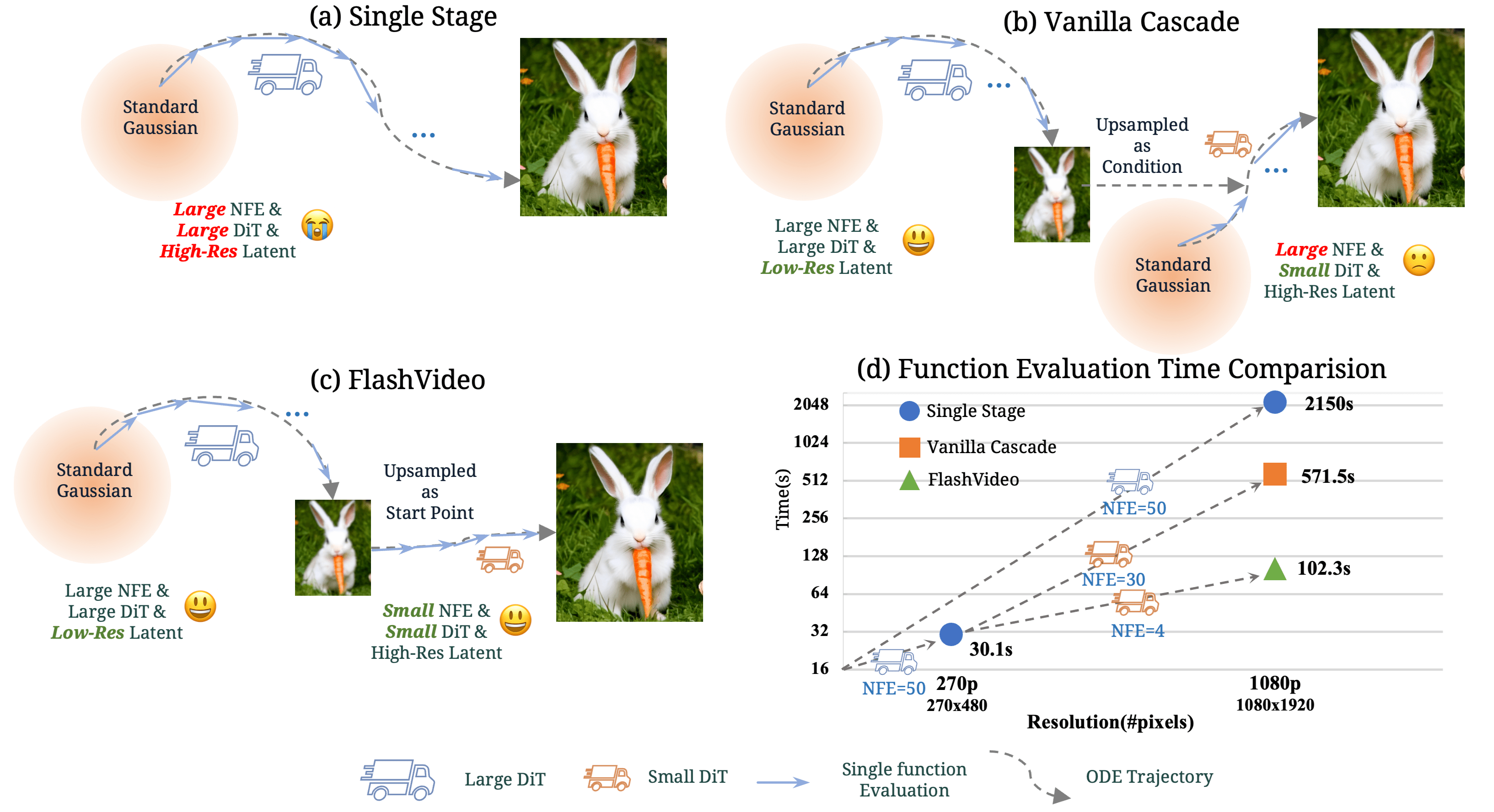}
\end{center}
\caption{ \textbf{Comparison between FlashVideo and other text-to-video generation paradigms.} 
(a) Single Stage DiT suffers from an explosive increase in computation cost when generating at large resolutions, rising from 30s to 2150s (\textcolor{blue}{circle} in (d)) when increasing the resolution from 270p to 1080p. (b) Though the vanilla cascade can reduce the model size in the high resolution, its second stage still samples from Gaussian noise and only uses the first-stage results as a condition. This approach cannot effectively reduce the number of function evaluations at high resolution and still costs 571.5s ({\textcolor{earthyellow}{square} in (d)}) to generate a 1080p video. (c) In contrast, FlashVideo not only decreases the model size in the second stage but also starts sampling from the first-stage results,  requiring only 4 function evaluations at high resolution while integrating a wealth of visually pleasant details, which can generate 1080P video with only 102.3s (\textcolor{green}{triangle} in (d)). Details on obtaining these statistics are provided in our \supp. }
\end{figure}\label{fig:teaser}

These requirements arise from the need to tackle key challenges in video generation, particularly ensuring high prompt fidelity and visual quality. First, achieving fidelity in both content and motion demands the model to encode extensive world knowledge. Research has shown significant improvements when increasing model parameters ($C$) from 2 billion to 12 billion~\citep{cogvideox, kong2024hunyuanvideo}. Additionally, an adequate number of denoising steps ($N$)~\citep{moviegen, kong2024hunyuanvideo, cogvideox} is essential for generating high-quality videos. While some efforts to reduce the number of steps have shown promising progress~\citep{ding2024dollar}, they are limited to lower resolutions and simpler motions. Moreover, visual quality has been proven to be tightly tied to resolution in text-to-image generation  ($H \times W$)~\citep{blattmann2023align, chen2025pixart, ren2024ultrapixel}, and for T2V tasks, the integrity of motion ($T$) must also be maintained. However, the combination of these challenges—large parameters, sufficient denoising steps, and high resolution—significantly increases the computational cost. For instance, a 5-billion-parameter model takes 2150s to generate 1080p videos, up from just 30s at the 270p resolution (Figure~\ref{fig:teaser}~(d)). 

To overcome these challenges, we introduce FlashVideo, a two-stage framework designed to separately optimize prompt fidelity and visual quality, as illustrated in Figure~\ref{fig:teaser}~(c). In the first stage, we focus on generating video content and motion that closely aligns with the user prompt. By operating at a lower resolution (\textit{e.g.}, 270p), even though we utilize a large model with 5 billion parameters with 50 evaluation steps, the model still remains efficient, requiring only 30 seconds function evaluation times (as shown in Figure~\ref{fig:teaser}~(d)). And as demonstrated in our experiments (Section.~\ref{exp:low_res}), this approach preserves semantic fidelity and motion smoothness. In the second stage, we enhance the generated video at 1080p, focusing on fine-grained detail enhancement while minimizing computational overhead. This is achieved using a lighter 2-billion-parameter model and an efficient flow-matching process with fewer evaluation steps. The two-stage framework effectively balances computational efficiency with high-quality results. 

While previous two-stage frameworks~\citep{upscaleavideo, lavie, venhancer} treat the first-stage low-resolution output as a condition and begin the second stage from Gaussian noise (Figure~\ref{fig:teaser}~(c)), this design requires 30–50 evaluation steps and still incurs significant computational cost (\textit{e.g.}, 571 seconds for 1080p generation). In contrast, FlashVideo uses flow matching to directly traverse ODE trajectories from first stage low-quality video to the final high-quality videos, eliminating the need to start from Gaussian noise. The flow matching target also tries to constrain the ODE trajectories to be straight. This design efficiently reduces the number of function evaluations to just 4 steps. As a result, FlashVideo reduces the function evaluation time for 1080p videos to just 102s, nearly $\sfrac{1}{20}$ of the time required by a single-stage model (Figure~\ref{fig:teaser}~(a)), and 5 times faster than vanilla cascade frameworks (Figure~\ref{fig:teaser}~(b)).

Despite this speedup, crafting an appropriate degradation simulation remains essential: naive pixel‑space operations (\textit{e.g.}, resizing and blurring~\citep{RealBasicVSR, yang2023mgldvsr}) fail to reproduce the fine‑scale artifacts introduced by model synthesis. To this end, we inject noise in the latent representation to eliminate all small structures in real videos, forcing their faithful reconstruction in the second stage. Moreover, we demonstrate that—within 3D full‑attention architectures, which are critical for spatio‑temporal coherence—the degradation intensity must be carefully tuned to prevent the model from ``cheating'' by merely copying details across frames, a pitfall neglected in prior work~\citep{stablecascade,supir,lavie,venhancer}.

In summary, our contributions are:

\begin{itemize}[leftmargin=*] 
\item We propose FlashVideo, a method that decouples video generation into two objectives: prompt fidelity and visual quality. By tailoring model sizes, resolutions, and optimization strategies in two stages, our approach achieves superior effectiveness and efficiency compared to existing methods.
\item  Innovatively, we construct nearly straight ODE trajectories starting from low-quality videos to high-quality videos through flow matching, which enables ample detail to be integrated into the video within only 4 function evaluations.
\item Our method achieves top-tier performance on VBench-Long (83.29 score) while achieving impressive function evaluation time. The two-stage design allows users to preview initial output before full-resolution generation, curtailing computational costs and wait times.
\end{itemize}

\section{Related Work}

\paragraph{Video generation models.} Recent advancements in text-to-video (T2V) generation have been remarkable~\citep{yan2021videogpt, cogvideo, videopoet, ho2022video, blattmann2023align, blattmann2023stable, sora, kling, vidu, luma2024dream, moviegen, pyramid_flow}. Key breakthroughs have been driven by the introduction of video diffusion and flow-matching algorithms~\citep{diffusion, ho2020denoising, rectifiedflow, fm}, alongside scaled text-video datasets and DiT parameters~\citep{peebles2023scalable}. Despite impressive generation quality, a major challenge remains the high computational cost, particularly for generating high-resolution videos.

\paragraph{Cascade diffusion models.} Numerous attempts have been made to explore cascade architectures in the text-to-image and text-to-video domains~\citep{saharia2022image, gu2023matryoshka, ho2022cascaded, stablecascade, upscaleavideo, supir, lavie, venhancer}. Researchers are motivated by the challenge that generating high-resolution images/videos in a single stage is both difficult and resource-intensive. In a cascade design, generation starts with a low-resolution sample, followed by an upsampling model to enhance visual appeal at higher resolutions. However, most methods perform the second-stage upsampling from pure noise, conditioning it on the low-resolution input, which requires a large number of function evaluations. While~\citep{zheng2024cogview3, teng2023relay, i2vgen, xing2024simda} have attempted to start from the first-stage distribution, their theories and implementations are complex, resulting in a high number of inference steps. Moreover,~\citep{boosting} proposes a pure super-resolution method for T2I using flow matching, but the limited generative priors in the second-stage model hinder substantial visual improvements. In this paper, we adhere to the principle of retaining only the most effective designs, developing FlashVideo, an efficient yet simple two-stage framework that achieves high-quality, high-resolution video generation with excellent computational efficiency.

\paragraph{Diffusion speeding up.} The generation process in diffusion models can be viewed as solving ordinary differential equations. To reduce the number of function evaluations, researchers have developed advanced samplers~\citep{song2020denoising, lu2022dpm, zhang2022fast}. Additionally, techniques for distilling pre-trained diffusion models into fewer steps have shown success~\citep{salimans2022progressive, meng2023distillation, yin2024one, nguyen2024swiftbrush, berthelot2023tract}. Adversarial training has also been employed to create few-step generators~\citep{xu2024ufogen, sauer2025adversarial, lin2024sdxl}. Recently, rectified flow~\citep{rectifiedflow} with straight ODE trajectories has been introduced, further refined by subsequent works~\citep{liu2023instaflow, yan_perflow_2024}, to enable faster sampling in T2I. However, few attempts have been made in the T2V field, where the added time dimension complicates the trajectories and increases computational demands. While some efforts to reduce the number of steps in T2V have shown promise~\citep{ding2024dollar}, they remain limited to low resolutions and simple motion. In this work, we propose an efficient flow matching pipeline that enables high-resolution video generation. Notably, the acceleration techniques discussed above are compatible with our framework, allowing for further speed improvements in both stages.

\section{Method}
\label{exp:method}

\begin{figure}[t]
\begin{center}
\includegraphics[width=1\linewidth]{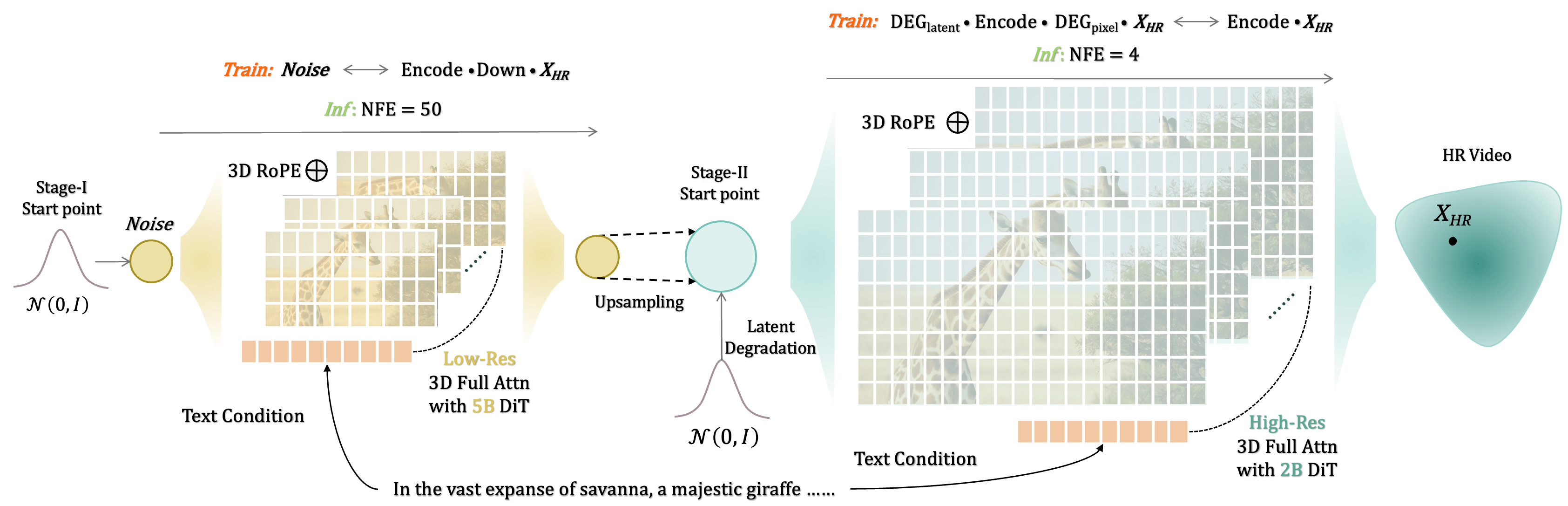}
\end{center}
\caption{ \textbf{The overall pipeline of FlashVideo}. FlashVideo adopts a cascade paradigm comprised of a 5-billion-parameter DiT at the low resolution (\textit{i.e.}, Stage \Romannum{1}) and a 2-billion-parameter DiT at a higher resolution (\textit{i.e.}, Stage \Romannum{2}). The 3D RoPE is employed at both stages to model the global and relative spatiotemporal distances efficiently.  We construct training data pairs for Stage \Romannum{1} by randomly sampling Gaussian noise and low-resolution video latent. For Stage \Romannum{2}, we apply both pixel and latent degradation to high-quality videos to obtain low-quality latent values. These are then paired with high-quality latents to serve as training data. During inference, we retain a sufficient $NFE=50$ at a low resolution of 270p  for Stage \Romannum{1}. The generated videos retains high fidelity and seamless motion, albeit with detail loss. These videos are then upscaled to a higher resolution of 1080p and processed by latent degradation. With only 4 steps, our Stage \Romannum{2} regenerates accurate structures and rich high-frequency details.
}
\label{fig:method}
\end{figure}

\subsection{Overview}

In the FlashVideo framework, video pixels $x \in \mathbb{R}^{H \times W \times T}$ are first compressed into latent features $f \in \mathbb{Q}^{h \times w \times t}$ using a 3D causal VAE~\citep{cogvideox}, where $h = \sfrac{H}{8}$, $w = \sfrac{W}{8}$, and $t = \sfrac{(T - 1)}{4} + 1$. The model is designed to generate 6-second videos (with 8 frames per second, so \( T = 49 \)) at 1080p resolution. As shown in Figure~\ref{fig:method}, we then employ a two-stage, low-to-high-resolution generation pipeline, where each stage is optimized with tailored model sizes and training strategies to ensure computational efficiency. The following subsections provide a detailed description of each stage.

% \subsection{Stage-\Romannum{1}}
\subsection{Low-Resolution Stage \Romannum{1}}

In the first stage, the goal is to generate videos with well-aligned content and motion corresponding to the input prompt. To achieve this, we initialize with a large-capacity model, CogVideoX-5B~\citep{cogvideox}, which contains 5 billion parameters. For improved computational efficiency, we perform parameter-efficient fine-tuning (PEFT) to adapt the model to a lower resolution of 270p. We find that adjusting the target resolution of the MMDiT architecture~\citep{sd3} is straightforward, which is achieved by applying LoRA~\citep{lora} with rank 128 to all attention~\citep{attention}, FFN, and adaptive layer normalization~\citep{adaln} layers. Compared to full-parameter tuning, PEFT demonstrates greater robustness, especially when fine-tuned with a small batch size of 32. In contrast, full-parameter tuning with such a small batch size significantly degrades generation quality. All other configuration settings, including the denoising scheduler and  prediction target, are kept consistent with CogVideoX-5B.

\subsection{High-Resolution Stage \Romannum{2}}
\label{exp:stage2}

\paragraph{Model architecture.} For fine-grained detail enhancement, we employ another model that adheres to the block design specified in CogvideoX-2B~\citep{cogvideox}. But, we replace the original position frequency embedding with 3D RoPE~\citep{rope}, as it offers better scalability for higher resolutions during inference (see Figure~\ref{fig:rope}).  Unlike the approach in~\citep{venhancer}, which uses spatial-temporal decomposition and time-slicing attention, we find that utilizing full 3D attention is crucial for maintaining consistency of enhanced visual details in videos with significant motion and scale variance, as shown in Figure~\ref{fig:comp_consis} and discussed in Section~\ref{sec:3d_attn}. As illustrated in Figure~\ref{fig:method}, the language embedding from the first stage is directly utilized in this stage.

\paragraph{Low-cost resolution transport.} Applying the conventional diffusion process at the high-resolution stage—starting from Gaussian noise and conditioned on low-resolution video—demands substantial computational resources. To improve efficiency while maintaining high-quality detail generation, we adopt flow matching~\citep{rectifiedflow, fm} to map the low-resolution latent representation, $\mathbf{Z}_{LR}$, to the high-resolution latent representation, $\mathbf{Z}_{HR}$. Intermediate points are computed through linear interpolation between $\mathbf{Z}_{LR}$ and $\mathbf{Z}_{HR}$, as outlined in Algorithm~\ref{alg:train}. This approach eliminates redundant sampling steps at the initialization phase and avoids reliance on additional control parameters, such as those proposed in~\citep{controlnet, supir, venhancer}. Furthermore, the $t$-independent target $\mathbf{Z}_{HR} - \mathbf{Z}_{LR}$ results in straighter ODE trajectories, enabling few-step generation. During training, $\mathbf{Z}_{LR}$ is simulated, as discussed later. In the testing phase, noise-augmented videos generated in the first stage serve as the starting point, and a commonly used Euler solver with $S=4$ steps, as outlined in Algorithm~\ref{alg:inf}, is employed. Other higher-order solvers can also be used for practical applications.

\begin{center}

\begin{minipage}[t]{0.5\linewidth}

  \centering
  \scalebox{0.86}
  {
  \begin{algorithm}[H]
    \caption{\small{~Training Stage}} \label{alg:train}
    \small{
    \textbf{Input: } High quality video dataset $D_{HR}$, model $F_\theta$ with parameters $\theta$, VAE encoder $\mathcal{E}$ \\
    \textbf{Procedure: }\\
    $\:\:$ \textbf{Repeat} \\
    $\:\:\:\: \mathbf{X}_{HR} \sim \mathbf{D}_{HR}$ \\
    $\:\:\:\: \mathbf{Z}_{HR} = \mathcal{E}(\mathbf{X}_{HR})$ \\
    $\:\:\:\: \mathbf{Z}_{LR} = DEG_{latent}(\mathcal{E} (DEG_{pixel}(\mathbf{X}_{HR})))$
    
    $\:\:\:\: Target =  \mathbf{Z}_{HR} - \mathbf{Z}_{LR}$ \\
    $\:\:\:\: t \sim  Uniform([0,1])$ \\
    $\:\:\:\: \mathbf{Z}_{t} = (1 - t) \cdot \mathbf{Z}_{LR} + t \cdot \mathbf{Z}_{HR}$ \\
    $\:\:\:\:$ Take gradient descent step on \\
    $\:\:\:\:\:\:\:\: \nabla_\theta \left\| Target - F_\theta\left( \mathbf{Z}_{t}, t \right) \right\|^2 $

    $\:\:$ \textbf{Until} Converged \\
    \textbf{Return:} Model $F_\theta$
    }
  \end{algorithm}
  }
\end{minipage}%
\begin{minipage}[t]{0.5\linewidth}
  % \vspace{0pt}
  \centering
  \scalebox{0.81}
  {
  \begin{algorithm}[H]
    \caption{\small{~Inference Stage}} \label{alg:inf}
    \small{
    \textbf{Inputs: } The video sample $\mathbf{X}_{LR}$ generated during the first stage, model $F_\theta$ with parameters $\theta$, VAE encoder $\mathcal{E}$ and VAE decoder $\mathcal{D}$, step number $S$  \\
    \textbf{Procedure: }\\

    $\:\: \mathbf{Z}_{LR} = DEG_{latent}(\mathcal{E}(\mathbf{X}_{LR})))$ \\
    $\:\: \Delta_{t} = \sfrac{1}{S} $ \\
    $\:\:  Z =  \mathbf{Z}_{LQ} $ \\
    $\:\:  t = 0 $ \\
    $\:\:$ \For {$step~in~[0, 1, \cdots, S-1 ]$}
    {
        $\Delta_z = F_\theta\left(Z, t\right) * \Delta_{t}$ \\
        $Z = Z + \Delta_z$ \\
        $t = t + \Delta_t$ \\
    }
     $\:\: \mathbf{Z}_{HR} = Z$ \\
     $\:\: \mathbf{X}_{HR} = \mathcal{D}(\mathbf{Z}_{HR})$ \\
    \textbf{Return: } High quality video $\mathbf{X}_{HR}$
    }
  \end{algorithm}
  }
\end{minipage}
\end{center}

\label{exp:simulation} \paragraph{Low quality video simulation.}

\begin{figure}[!t]
\begin{center}
\includegraphics[width=1\linewidth]{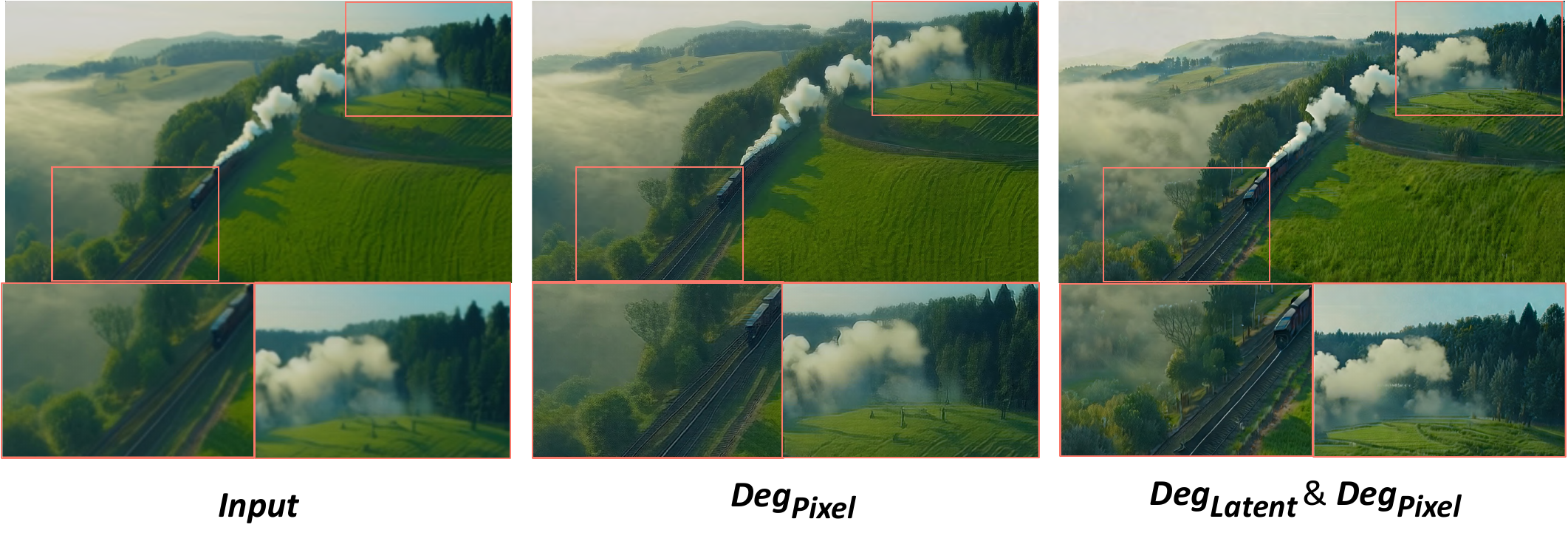}
\end{center}

\caption{\textbf{Visual showcase of $DEG_{pixel}$ and $DEG_{latent}$ impact on quality enhancement.}  From left to right, the first is the $input$, generated by the first-stage model. The term $DEG_{pixel}$ stands for the improved result yielded from the model trained only with pixel-space degradation, which adds high-frequency details to the $input$. Further, $DEG_{pixel}$ \& $DEG_{latent}$ refers to the enhanced result with model trained under both types of degradation, which further  improves small structures, such as generating branches for small trees. The improvement is significantly apparent when compared to pixel degradation only.}

\label{fig:deg_ab}
\end{figure}

To train the second-stage model, we establish paired low-resolution and high-resolution latent representations, $\mathbf{Z}_{LR}$ and $\mathbf{Z}_{HR}$. Starting from a high-quality video $\mathbf{X}_{HR}$, we apply a sequence of blur and resize operations with randomized strengths in the pixel space (details provided in the \supp), yielding the low-resolution video. This process, denoted as $DEG_{pixel}$, is outlined in Algorithm~\ref{alg:train}. Training on this simulated data enables the model to enhance images with high-frequency details, improving overall clarity, as demonstrated in Figure~\ref{fig:deg_ab}.

However, simulating low-resolution data solely through $DEG_{pixel}$ retains strong fidelity between low- and high-resolution videos, which limits the model's ability to regenerate accurate structures for small objects at high resolutions—especially when artifacts are present in the first-stage output. This limitation often manifests when there are poor structural representations for small objects, such as blurry tree branches in Figure~\ref{fig:deg_ab} or distorted eye features in Figure~\ref{fig:stage2} (e). To address this issue, we introduce latent degradation, $DEG_{latent}$, which perturbs the latent representation with Gaussian noise. This approach allows the model to diverge from the input and generate more reasonable structures for small objects. As shown in Figure~\ref{fig:deg_ab}, compared to $DEG_{pixel}$, the combination of $DEG_{latent}$ enables the model to produce sharper and more detailed tree branches and tiny background objects, significantly enhancing visual quality.

The overall simulation process during training can be described as follows: First, pixel-space degradation is applied to the high-quality video, yielding a degraded version. This is then encoded into the latent space, represented as:
\begin{equation}
Z = \mathcal{E} \left( DEG_{pixel} \left (\mathbf{X}_{HR} \right) \right) \,.
\end{equation}
Next, the latent representation is blended with Gaussian noise $ n \sim N(0,1) $ to simulate low-quality latents, defined as:
\begin{equation}\label{eq:lq-formulation} 
Z_{LR} = DEG_{latent}(Z) = \alpha_{step} \cdot Z + \beta_{step} \cdot n \,, \quad \text{where } \alpha_{step}^2 + \beta_{step}^2 = 1 \,.
\end{equation}
The parameter $step$ determines the strength of noise augmentation. To ensure the model can perceive the noise strength in the latent space, we introduce a noise strength embedding, which is added to the time embedding. At the inference stage, only $DEG_{latent}$ is applied to the first-stage output.   However, setting this degradation strength requires careful design in our full attention framework, as discussed in the subsequent paragraph.

\paragraph{Latent degradation strength search.} \label{exp:simulation}  Although the Stage II full‐attention architecture enhances spatio‐temporal coherence (see Figure~\ref{fig:comp_consis}), it may exhibit a ``cheating'' behavior: under weak latent degradation (\textit{e.g.}, noise steps $<$ 400), the model simply copies details across frames, leaving artifacts uncorrected. Conversely, overly aggressive degradation (\textit{e.g.}, noise steps $> 800$) overwhelms the generator, degrading overall fidelity. Hence, precise tuning of the latent degradation magnitude ($DEG_{\text{latent}}$) is essential to erase undesirable artifacts while retaining structures produced in Stage I, which is largely unexplored. Traditional super‐resolution approaches \citep{RealBasicVSR, yang2023mgldvsr} do not target model‐introduced artifacts (Figure~\ref{fig:comp_frame}), while recent diffusion-based video enhancement techniques without global attention~\citep{lavie, upscaleavideo, venhancer} avoid this cheating at the expense of temporal consistency (Figure~\ref{fig:comp_consis}). To identify an optimal $DEG_{\text{latent}}$ range, we begin training with a broad noise‐step interval (600–900), then refine it to 650–750 based on empirical performance (Table~\ref{table:latent_deg}).

\paragraph{Coarse-to-fine training.}

Training directly on high resolution requires substantial computational costs. The use of 3D RoPE~\citep{rope, cogvideox}, a relative spatiotemporal encoding, offers good resolution scalability for our model (Section~\ref{sec:pos_emb}). As a result, we first conduct large-scale pre-training on low-resolution images and videos ($540 \times 960$) before extending to the target resolution of 1080p ($1080 \times 1920$). Observing obvious performance fluctuations in the later stages, we further fine-tune the model with a small set of high-quality samples aligned to human preferences. This low-cost additional fine-tuning stage greatly improves the model's performance.

\section{Experiments}

\subsection{Data Collection}

We construct a high-quality dataset by first collecting a large corpus of 1080p videos, followed by aesthetic and motion-based filtering, resulting in 2 million high-quality samples. Motion filtering is performed using RAFT~\citep{teed2020raft} to compute the average optical flow, discarding clips with low motion scores \((<1.1)\). To ensure the second-stage model learns diverse texture details, we further collect 1.5 million high-quality images at a resolution of $2048 \times 2048$. All videos and images are annotated with detailed captions generated by an internal captioning model. For human preference alignment, we manually curate a subset of 50,000 videos exhibiting high aesthetic quality, rich textures, and significant motion diversity.

\subsection{Training Setup}
\label{sec:training}

For training the first-stage model, we use only video data, which are resized to the 270p resolution. The model is trained for 50,000 iterations with a batch size of 32 and a base learning rate of $4 \times 10^{-5}$. We employ the AdamW optimizer with $\beta_1 = 0.9$, $\beta_2 = 0.95$, a weight decay of $1 \times 10^{-4}$, and gradient clipping set to 0.1.

\label{exp:optimizer}

The second-stage model, which includes both pre-training and human preference alignment, is trained with a batch size of 64, while other hyperparameters remain consistent with those used in the first stage. The pre-training is structured into three phases: (1) training for 25,000 iterations on $540 \times 960$ image patches cropped from $2048 \times 2048$ high-resolution images, (2) 30,000 iterations on a mixed dataset of $540 \times 960$ image patches and videos at a 1:2 ratio, and (3) training on full-resolution  $1080 \times 1920$ videos for 5000 iterations. Finally, we perform (4) fine-tuning on the human preference alignment dataset for 700 iterations. For latent degradation, we initially apply noise within the step range of 600–900 for phases (1), (2), and the first 1000 iterations of (3). Based on the findings in Table~\ref{table:latent_deg}, we then narrow the noise range to 650–750 for the remaining training in (3) and (4).

\subsection{Qualitative Results}
In this section, we present visualizations of the two-stage video generation results based on various user prompts. The first-stage output prioritizes high fidelity in both content and motion, while the second stage further refines details and mitigates generation artifacts, thereby enhancing overall visual quality.

\begin{figure}[!t]
\begin{center}
\includegraphics[width=1\linewidth]{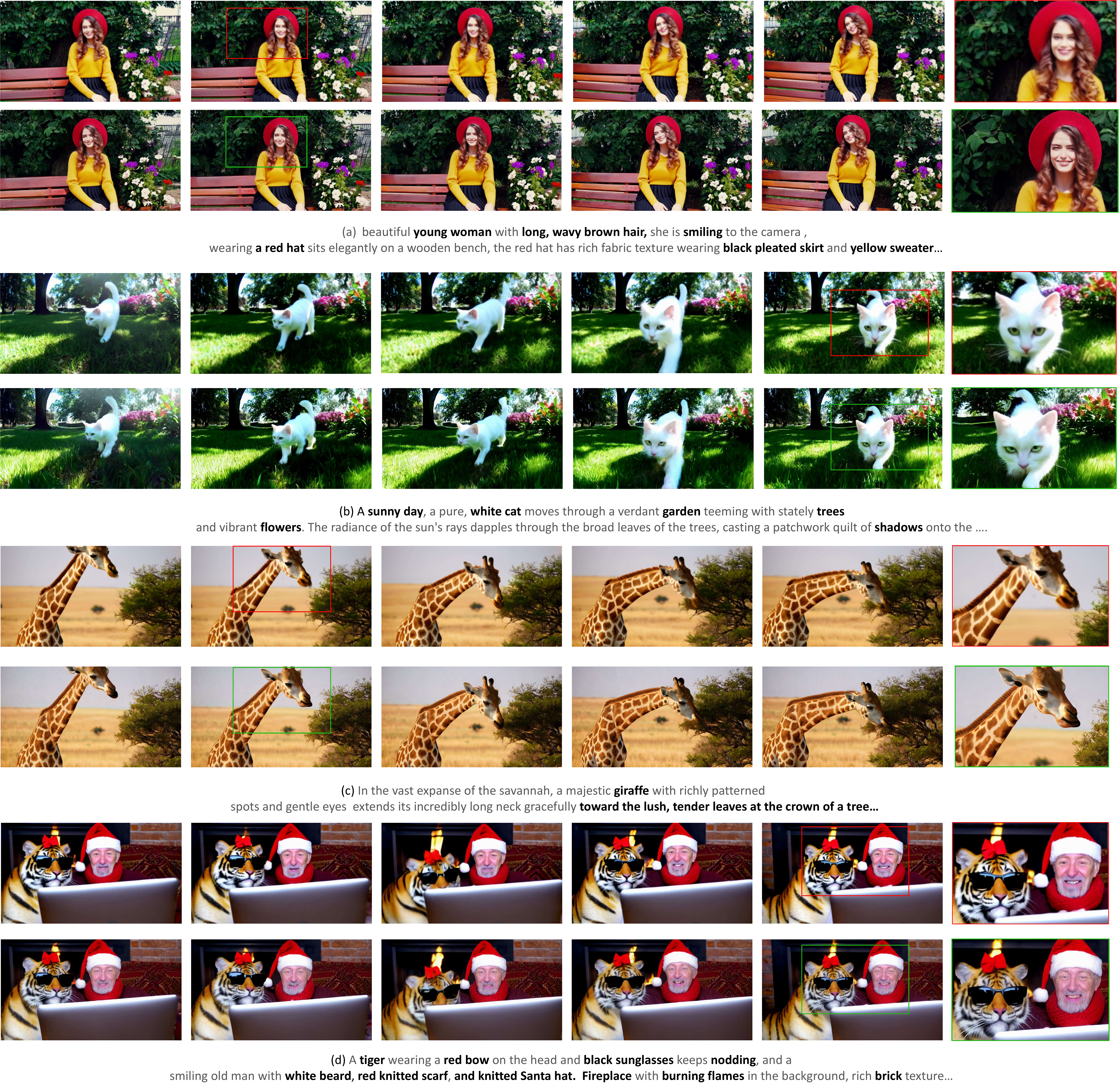}
\end{center}
\caption{\textbf{Generated videos of FlashVideo.} The results in the top and bottom rows are from Stage \Romannum{1} and Stage \Romannum{2}, respectively. Stage \Romannum{1} generates videos with natural motion and high prompt fidelity, as evident from the visual elements (\textbf{bold} in prompts). However, they lack detailed structures for small objects and high-frequency textures (see the \textcolor{red}{red} box). In Stage \Romannum{2}, details are significantly enriched (see the \textcolor{green}{green} box), while content remains highly consistent with the original. Visualization results are compressed. More uncompressed cases can be found on our \href{https://jshilong.github.io/flashvideo-page/}{project page}.}\label{fig:demo}
\end{figure}

\begin{figure}[!t]
\begin{center}
\includegraphics[width=1\linewidth]{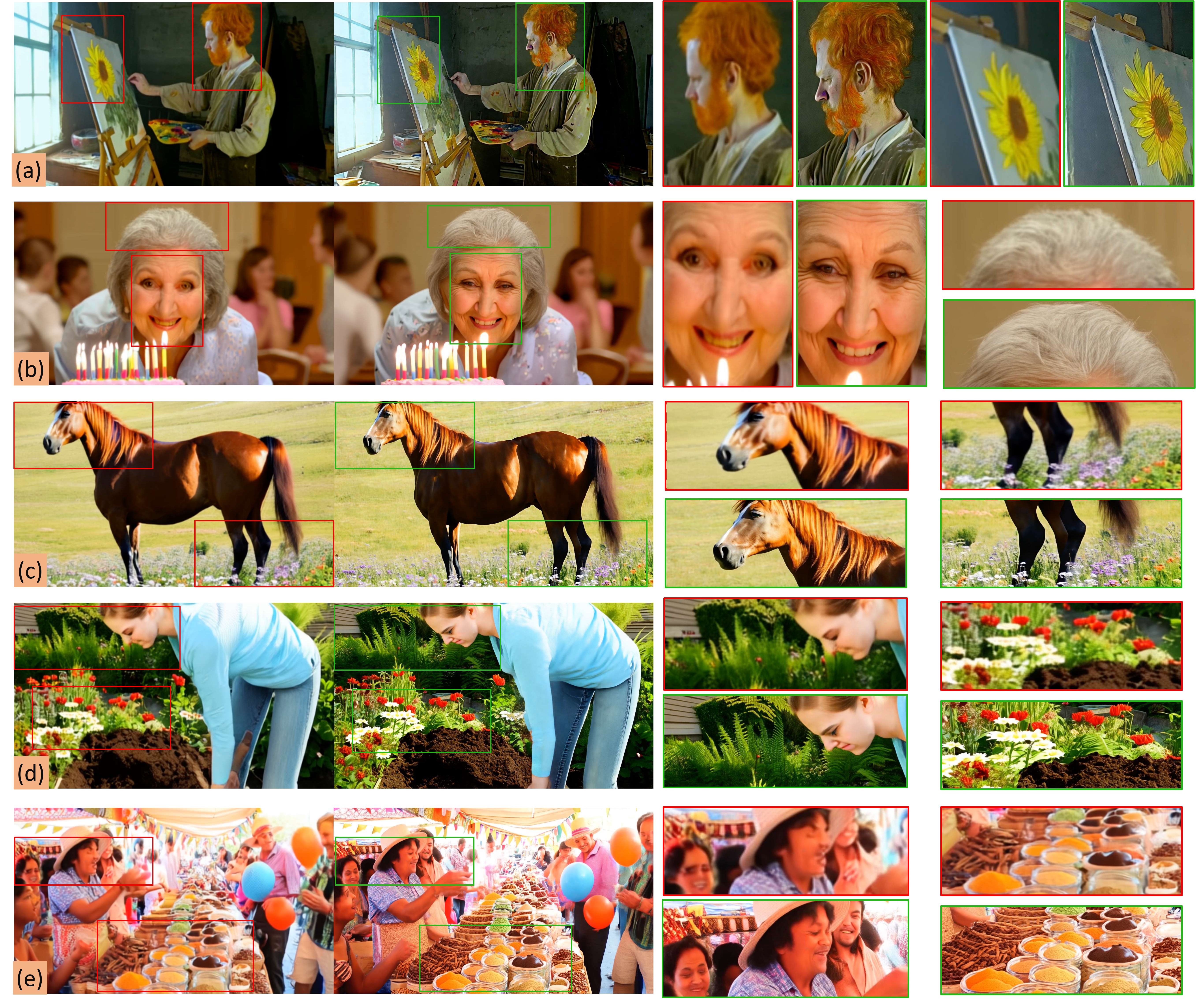}
\end{center}
\caption{\textbf{Quality improvements in Stage \Romannum{2} }. We mark regions with artifacts and lacking detail in the first-stage videos using \textcolor{red}{red} boxes, while improvements from the second stage are highlighted in \textcolor{green}{green}. Zoom in for a better view. Our Stage \Romannum{2} significantly elevates visual quality across diverse content—enhancing oil painting–style sunflowers in (a), refining wrinkles and hair in (b), enriching texture structures of animals and plants in (c) and (d), and mitigating facial and object artifacts in (e).
}
\label{fig:stage2}
\end{figure}

\paragraph{Two-stage generation results.}
As shown in Figure~\ref{fig:demo}, the first-stage outputs (top rows) exhibit strong prompt fidelity with smooth motion. The key visual elements specified in the prompt, highlighted in \textbf{bold}, are accurately generated. However, artifacts and insufficient texture details, marked by the red bounding box, may still be present. In contrast, the second-stage outputs (bottom rows) significantly improve visual quality by refining small objects with plausible structures and enhancing texture richness. Notable improvements include the refined depiction of human faces (a, d), the detailed rendering of animal fur (b, c), the intricate structures of plants (a, b), and the enhanced fabric textures (d), as highlighted in the green bounding box of the second row. Moreover, despite substantial motion, high-frequency details remain temporally consistent, owing to the full attention mechanism integrated into the second stage. More uncompressed cases can be found on our \href{https://jshilong.github.io/flashvideo-page/}{project page}.

\paragraph{Artifact correction and detail enhancement in Stage \Romannum{2}.}
To further demonstrate the effectiveness of the second-stage refinement, we provide additional examples of key frames in Figure~\ref{fig:stage2}. Compared to the first-stage outputs (marked in red), the second-stage results (marked in green) exhibit significant improvements by suppressing artifacts and enriching fine details. These enhancements are evident in the more coherent depiction of oil painting-style sunflowers in (a), the refined rendering of wrinkles and hair in (b), the improved texture structures of animals and plants in (c) and (d), and the correction of facial and object artifacts in (e).

\subsection{Quantitative Results}

We first evaluate our model on the VBench-Long~\citep{huang2024vbench} benchmark utilizing  its long prompt. Subsequently, we assess the visual quality improvements achieved in Stage \Romannum{2} by employing several widely used non-reference image and video quality assessment metrics.

\begin{table*}[t]
\centering
%\scriptsize
\tiny
\setlength{\tabcolsep}{1.5pt}
\renewcommand{\arraystretch}{1.0}
\begin{tabular}{l|l|ll|llllllllllllllll}
     Method  & 
    \makecell[bc]{\rotatebox{75}{Total Score}} & 
    \makecell[bc]{\rotatebox{75}{Quality Score}} & 
    \makecell[bc]{\rotatebox{75}{Semantic Score}} & 
    \makecell[bc]{\rotatebox{75}{subject consistency}} & 
    \makecell[bc]{\rotatebox{75}{background consistency}} & 
    \makecell[bc]{\rotatebox{75}{temporal flickering}} & 
    \makecell[bc]{\rotatebox{75}{motion smoothness}} & 
    \makecell[bc]{\rotatebox{75}{dynamic degree}} & 
    \makecell[bc]{\rotatebox{75}{aesthetic quality}} & 
    \makecell[bc]{\rotatebox{75}{imaging quality}} & 
    \makecell[bc]{\rotatebox{75}{object class}} & 
    \makecell[bc]{\rotatebox{75}{multiple objects}} & 
    \makecell[bc]{\rotatebox{75}{human action}} & 
    \makecell[bc]{\rotatebox{75}{color}} & 
    \makecell[bc]{\rotatebox{75}{spatial relationship}} & 
    \makecell[bc]{\rotatebox{75}{scene}} & 
    \makecell[bc]{\rotatebox{75}{appearance style}} & 
    \makecell[bc]{\rotatebox{75}{temporal style}} & 
    \makecell[bc]{\rotatebox{75}{overall consistency}} \\
    \toprule

HunyuanVideo & 83.24 & 85.09 & 75.82 &97.37 & 97.76 &99.44 & 98.99 & 70.83 & 60.36 & 67.56 & 86.10 & 68.55 &  94.40 & 91.60 & 68.68 & 53.88 & 19.80 & 23.89 & 26.44  \\
Vchitect(VEnhancer) & 82.24 & 83.54 & 77.06 & 96.83 & 96.66 & 98.57 & 98.98 & 63.89 & 60.41 & 65.35 & 86.61 & 68.84 & 97.20 & 87.04 & 57.55 & 56.57 & 23.73 & 25.01 & 27.57 \\
CogVideoX-1.5 & 82.17 & 82.78 & 79.76 & 96.87 & 97.35 & 98.88 & 98.31 & 50.93 & 62.79 & 65.02 & 87.47 & 69.65 & 97.20& 87.55& 80.25& 52.91& 24.89& 25.19& 27.30 \\
CogVideoX-5B & 81.61 & 82.75 & 77.04 & 96.23 & 96.52& 98.66& 96.92& 70.97 & 61.98 & 62.90& 85.23& 62.11& 99.40& 82.81& 66.35& 53.20& 24.91 & 25.38 & 27.59 \\
CogVideoX-2B & 80.91 & 82.18 & 75.83 & 96.78 & 96.63 & 98.89 & 99.02 & 59.86 & 60.82 & 61.68 & 83.37 & 62.63 & 98.00 & 79.41 &69.90& 51.14 & 24.80 & 24.36 & 26.66 \\
Mochi-1 & 80.13 &82.64 &70.08 & 96.99 & 97.28 & 99.40 & 99.02 & 61.85 & 56.94 & 60.64 & 86.51 & 50.47 & 94.60 & 79.73 & 69.24 & 36.99 & 20.33& 23.65 & 25.15 \\
LTX-Video & 80.00 & 82.30 & 70.79 & 96.56 & 97.20 & 99.34 & 98.96 & 54.35 & 59.81 & 60.28 & 83.45 & 45.43 & 92.80 & 81.45 & 65.43 & 51.07 & 21.47 & 22.62 & 25.19 \\
OpenSora-1.2 & 79.76 & 81.35 & 73.39 & 96.75 & 97.61 & 99.53 & 98.50 & 42.39 & 56.85 & 63.34 & 82.22 &51.83&91.20& 90.08 &68.56  &42.44  &23.95  &24.54  &26.85  \\
OpenSoraPlan-V1.1 &  78.00 & 80.91 & 66.38 & 95.73 & 96.73 & 99.03 & 98.28 & 47.72 & 56.85 & 62.28 & 76.30 & 40.35 & 86.80 & 89.19 & 53.11 & 27.17 & 22.90 & 23.87 & 26.52 \\
\midrule

FlashVideo$_{\scalebox{0.75}{$\scriptscriptstyle 8fps$}}$ & 82.80 & 82.99 & 82.03 & 96.91 & 96.77 & 98.56 & 96.84 & 63.47 & 62.55 & 66.96 & 90.02 & 81.47 & 99.00 & 85.71 & 83.20 & 55.34 & 24.64 & 25.23 & 27.65 \\
FlashVideo$_{\scalebox{0.75}{$\scriptscriptstyle 24fps$}}$ & 83.29 & 83.72 & 81.60 & 97.14 & 97.07 & 98.57 & 98.83 & 59.86 & 62.41 & 66.12  & 88.45 & 80.27 & 99.00 & 84.14 & 82.27 & 56.71  & 24.60  & 25.23 & 27.60 \\

\bottomrule 
\end{tabular}
\caption{\textbf{Comparison with state-of-the-art open-source models on VBench-Long benchmark~\citep{huang2024vbench}.}  This includes the recent HunyuanVideo~\citep{kong2024hunyuanvideo},   Vchitect-2.0 incorporated with VEnhancer~\citep{venhancer}, varying versions of CogVideoX~\citep{cogvideox}, Mochi-1~\citep{genmo2024mochi}, LTX-Video~\citep{HaCohen2024LTXVideo}, OpenSora~\citep{opensora} and OpenSoraPlan~\citep{lin2024open}. FlashVideo employs a cascade paradigm to deliver top-tier semantic fidelity and quality.}
\label{tab:vbench}
\end{table*}

\paragraph{VBench-Long benchmark.} We follow the standard evaluation protocol of VBench-Long, generating five videos per prompt. Noting that VBench metrics tend to favor higher frame rates, we apply a real-time video frame interpolation method~\citep{huang2022rife} to upscale the frame rate from 8 fps to 24 fps. This interpolation incurs negligible post-processing time (within 4 seconds), ensuring fair comparisons with high-frame-rate methods. A more detailed discussion on VBench’s frame rate preference is provided in the \supp.

\label{exp:low_res} As shown in Table~\ref{tab:vbench}, 
both our 8fps and 24fps models achieve high semantic scores exceeding 81. However, relying solely on the first-stage model results in aesthetic and imaging quality scores below top-tier methods, with 60.74 and 61.87 for 270p. After applying the second stage, both quality scores improve significantly, reaching state-of-the-art levels of approximately 62.55 and 66.96, respectively, as reported in Table~\ref{tab:vbench}. These results validate our approach of initially reducing the resolution in Stage \Romannum{1} to ensure high prompt fidelity at a lower computational cost, followed by quality enhancement in Stage \Romannum{2}. On the other hand, our entire functional evaluation only takes about 2 minutes, significantly outperforming other methods in terms of efficiency. For example, a concurrent work, Hunyuan Video~\citep{kong2024hunyuanvideo}, which achieves a total score of 83.24 using a larger 13B single-stage model, requires 1742 seconds for  function evaluation  to generate 720p $(720\times 1280)$ results. In contrast, our method not only demonstrates superior efficiency but also generates outputs at higher resolution. Furthermore, users can obtain preliminary previews in just 30 seconds for 270p, allowing them to decide whether to proceed with the second stage or refine the input prompt. This flexibility significantly enhances the user experience.

\begin{table*}[!t]
    \centering
    \scriptsize
    \setlength{\tabcolsep}{4.5pt}
    \renewcommand{\arraystretch}{1.0}
    \begin{tabular}{lcccccccc}
    \toprule
    &  & \multicolumn{4}{c}{\textbf{Frame Quality}} & \multicolumn{2}{c}{\textbf{Video  Quality}}  \\

    \cmidrule(lr){3-6}
\cmidrule(lr){7-8}
    
    & \#NFE / Time &  MUSIQ($\uparrow$) & MANIQA($\uparrow$) & CLIPIQA($\uparrow$) & NIQE($\downarrow$)   &  Technical($\uparrow$)   & Aesthetic($\uparrow$)    \\
\midrule

Stage \Romannum{1} ~(270p) &  50 / 30.1s  & 24.54   & 0.226 & 0.334 & 11.77 & 7.280 & 96.15  \\
Stage \Romannum{2}~(1080p) & 4 / 72.2s & \textbf{53.46}  & 
\textbf{0.302} & \textbf{0.436} & \textbf{5.380} & \textbf{11.68}  & \textbf{97.87}  \\

\bottomrule
\end{tabular}
\caption{Comparison of frame quality and video quality between two stages with Vbench-Long prompts.The best results are emphasized in \textbf{bold}.}
\label{table:two_stage}
\end{table*}

\begin{table*}[!h]
\centering
\resizebox{\linewidth}{!}{%
\setlength{\tabcolsep}{1.5pt}
\renewcommand{\arraystretch}{1.0}
\begin{tabular}{lcccccccccccc}
\toprule
&  & \multicolumn{4}{c}{\textbf{Frame Quality}} & \multicolumn{2}{c}{\textbf{Video  Quality}}  & \multicolumn{4}{c}{\textbf{Attributes}} \\
\cmidrule(lr){3-6}
\cmidrule(lr){7-8}
\cmidrule(lr){9-12}
    & \#NFE / Time &  MUSIQ($\uparrow$) & MANIQA($\uparrow$) & CLIPIQA($\uparrow$) & NIQE($\downarrow$)   &  Tech($\uparrow$)   & Aesth($\uparrow$) 
    & \makecell{First Stage\\as Preview}
    & \makecell{Artifacts\\Correction}
    & \makecell{Long-Range\\Detail\\Consistency}
    & \makecell{Avoids\\Detail\\Flickering} \\
\midrule
RealbasicVSR & 1 / 71.5s & \underline{54.26}  & 0.272 &\underline{0.418} &\underline{5.281}&  10.71 & \textbf{99.42} 
& {\textbf{$\checkmark$}} 
& {\textbf{$\times$}} 
& {\textbf{NA}} 
& {\textbf{$\checkmark$}} \\
Upscale-A-Video & 30 / 376.6s & 23.67 &0.201 & 0.285 & 12.02 & 7.690 & 97.61 
& {\textbf{$\checkmark$}} 
& {\textbf{$\times$}} 
& {\textbf{NA}} 
& {\textbf{$\times$}} \\
VEnhancer & 30 / 549.2s & 51.69  & \underline{0.280}& 0.385 &  5.330 & \underline{11.63} & 98.39 
& {\textbf{$\times$}} 
& {\textbf{$\checkmark$}} 
& {\textbf{$\times$}} 
& {\textbf{$\checkmark$}} \\
FlashVideo (Ours) &  4 / 72.2s  & \bf{58.69}    &   \textbf{0.296}   & \textbf{0.439}   &\textbf{4.501}  & \textbf{11.86}  & \underline{98.92} 
& {\textbf{$\checkmark$}} 
& {\textbf{$\checkmark$}} 
& {\textbf{$\checkmark$}} 
& {\textbf{$\checkmark$}} \\
\bottomrule
\end{tabular}
}

\caption{
\textbf{Comparison with video enhancement methods.}
Best results are \textbf{bold}, second-best \underline{underlined}. We highlight key design attributes for Stage II; only our tailored method meets all criteria.
Visual examples are available on the project page. “NA” indicates limited new details, preventing evaluation of long-range consistency.
}
\label{table:com_other}
\end{table*}

\paragraph{Frame and video quality assessment.} As shown in Table~\ref{table:two_stage}, we present a comprehensive comparison of visual quality between the two stages with all VBench-Long prompts. We utilize widely recognized image quality assessment metrics, including MUSIQ ($\uparrow$)~\citep{musiq}, MANIQA ($\uparrow$)~\cite{maniqa}, CLIPIQA ($\uparrow$)~\citep{clipiqa}, and NIQE ($\downarrow$)~\citep{niqe}, along with the video metric DOVER~\citep{dover}, to assess the perception of distortions (Technical $\uparrow$) and content preference and recommendation (Aesthetic $\uparrow$). It is evident that all metrics show significant improvements following the application of Stage \Romannum{2}. We argue that increasing the resolution in the second stage (Section~\ref{sec:pos_emb}), ultimately producing higher outputs (\textit{e.g.}, 2K), would further enhance visual quality, and this will be explored in future work.

\subsection{Comparison with Video Enhancement Methods}

To comprehensively evaluate the effectiveness of our tailored Stage \Romannum{2}, we compare it against several state-of-the-art video enhancement methods, including VEnhancer~\citep{venhancer}, Upscale-a-Video~\citep{upscaleavideo}, and RealBasicVSR~\citep{RealBasicVSR}. Our evaluation comprises both quantitative and qualitative analyses based on the first-stage outputs. Specifically, we construct a curated test set of 100 text prompts with detailed descriptions and generate the corresponding low-resolution 6-second 49-frame videos using Stage \Romannum{1}, incorporating diverse visual elements such as characters, animals, fabrics, and landscapes. We refer to this test set as Texture100. The following ablation study is also conducted on this test set.

 As shown in Table~\ref{table:com_other}, we summarize the key attributes for designing Stage II: maintaining high fidelity for a reliable preview, correcting artifacts from Stage I, and ensuring detail consistency across both adjacent frames (avoid flickering) and long-range frames. Notably, only \ours satisfies all these requirements.
The frame and video quality metrics are also reported in Table~\ref{table:com_other}, where FlashVideo consistently surpasses competing methods by a substantial margin while maintaining superior efficiency. Notably, although the GAN-based RealBasicVSR achieves competitive scores on some metrics, its outputs frequently exhibit excessive smoothing, indicating a misalignment between these metrics and human perceptual preferences. Consequently, we recommend interpreting quantitative evaluations as supplementary references while prioritizing qualitative assessments. On the other hand, the diffusion-based VEnhancer demonstrates stronger generative capabilities. However, its outputs often undergo significant deviations from the input, contradicting our core design principle of enhancing visual quality while preserving fidelity. Furthermore, VEnhancer employs separate spatial-temporal modules and time slicing instead of 3D full attention, leading to reduced content consistency across extended video sequences—an issue we will explore in subsequent discussions. Additionally, its high NFE results in increased computational overhead, making high-resolution generation time-intensive. In contrast, our model achieves nearly a sevenfold speedup over VEnhancer while producing sharper high-frequency details, as evidenced in Table~\ref{table:com_other}.

\begin{figure}[!t]
\begin{center}
\includegraphics[width=1\linewidth]{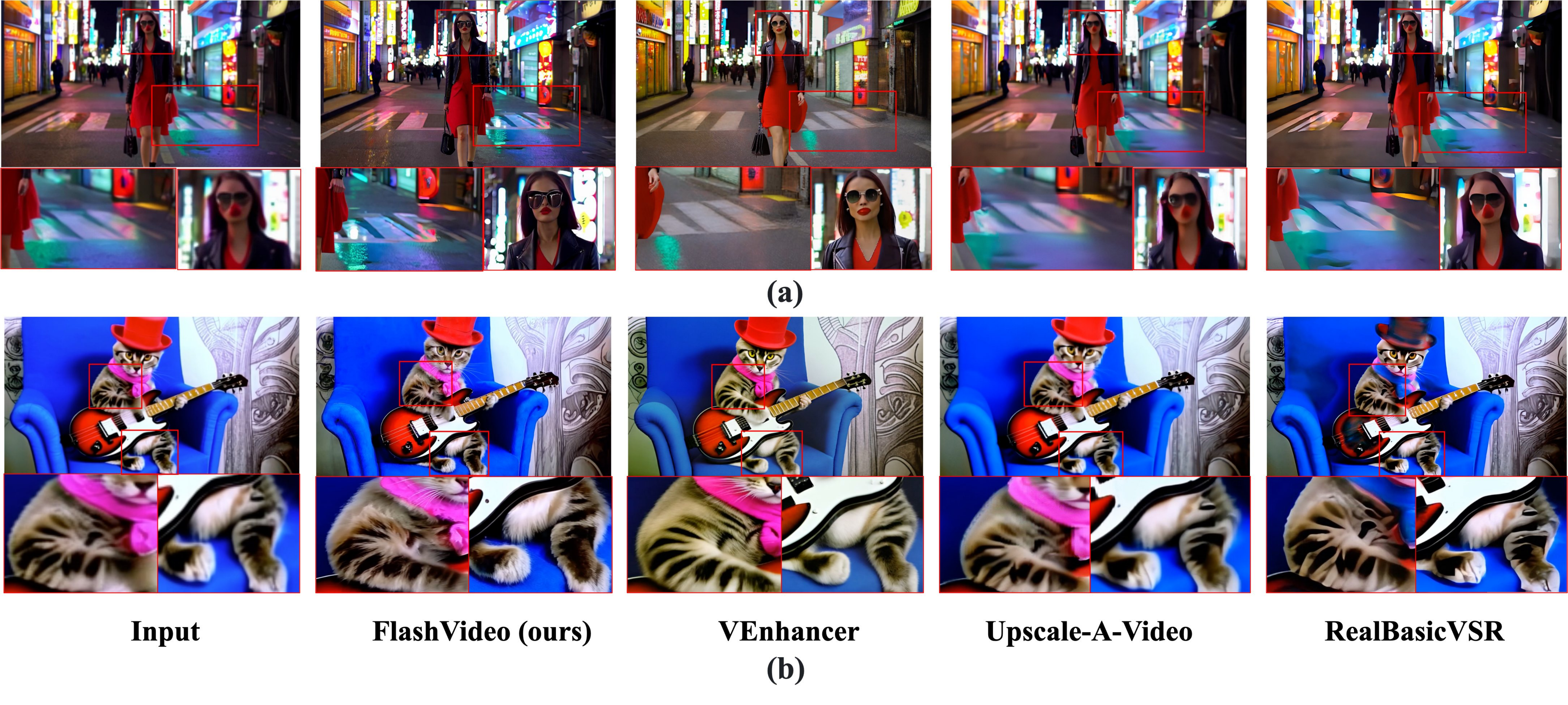}
\end{center}
\caption{\textbf{Visual comparison with various video enhancement methods}. We present our results alongside enhanced versions, derived from the first-stage outputs, of four video enhancement methods.}

\label{fig:comp_frame}
\end{figure}

\begin{figure}[!t]
\begin{center}
\includegraphics[width=1\linewidth]{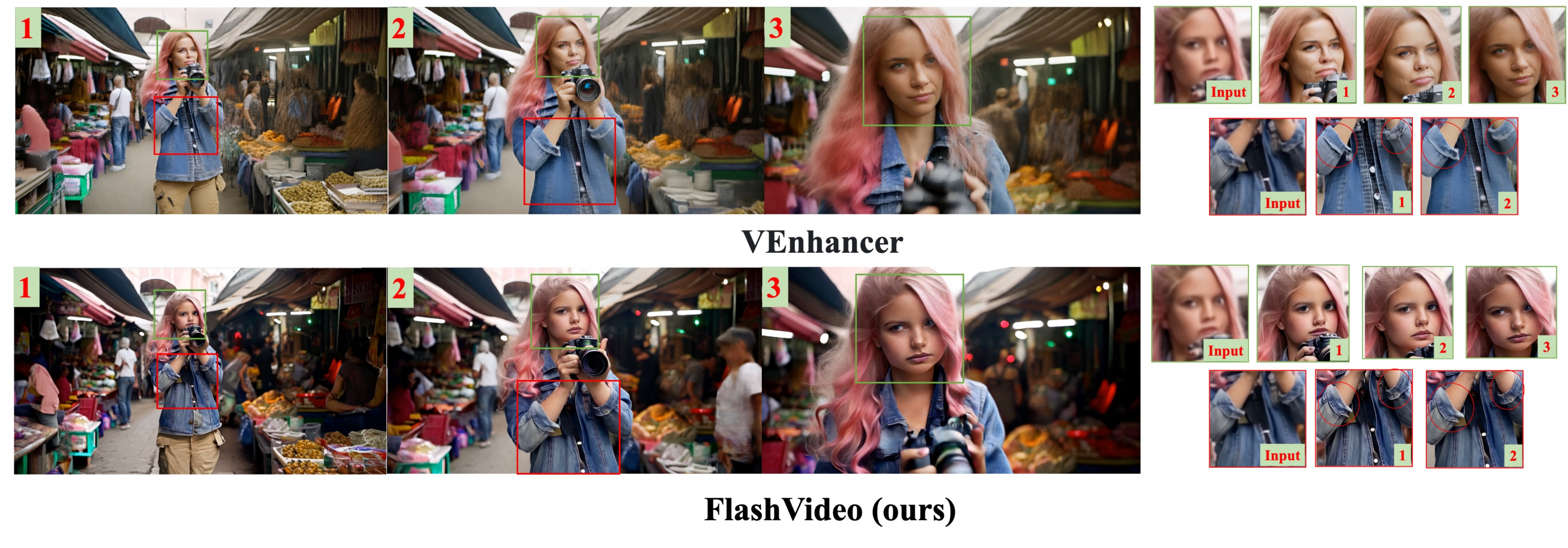}
\end{center}
\caption{\textbf{Comparison of long-range detail consistency in large-motion videos. }We select a first-stage generated video with significant motion and sample three key frames. The girl in this video undergoes substantial scale variation from distant to close-up views. VEhancer~\cite{venhancer}, with spatial-temporal module and time slicing, fails to preserve identity and detail consistency. In contrast, FlashVideo leverages 3D full attention to maintain consistent facial identity and texture details. 
}

\label{fig:comp_consis}
\end{figure}

Figure~\ref{fig:comp_frame} (a) illustrates a case where the woman's face contains noticeable artifacts, and the background appears blurry. Our method effectively reconstructs intricate facial details while enriching the background with high-frequency textures, maintaining both structural integrity and fidelity. 
In comparison, although VEnhancer yields a relatively clear face, it also significantly alters the background, losing fidelity entirely. Essential visual elements like ``standing water'' on the ground and the overall dim tones are completely lost. This result is contrary to our intent of using the first-stage results for preview. Other methods, such as Upscale-a-Video and RealBasicVSR, fail to correct facial artifacts and instead generate excessively smoothed patterns, further reducing realism. A similar trend is observed in Figure~\ref{fig:comp_frame} (b), where our approach delivers richer textures—such as distinct individual hairs on the cat’s body—while preserving consistency with the original input.
\label{sec:3d_attn}As discussed earlier, the full attention mechanism in our model plays a crucial role in maintaining content consistency, outperforming VEnhancer in this regard. Figure~\ref{fig:comp_consis} presents a sequence of three frames featuring substantial motion, where the camera transitions from a distant to a close-up view, leading to significant scale variations in the subject’s appearance. While both FlashVideo and VEnhancer exhibit clear improvements over the initial input, VEnhancer struggles to preserve facial identity across the key frames and introduces inconsistencies in fine details such as jacket textures and background elements. In contrast, our method effectively mitigates these issues, ensuring stable and coherent visual quality throughout the sequence.

\section{Ablation}

In this section, we conduct a series of ablation studies to evaluate the key designs of our approach. First, we examine the advantage of LoRA fine-tuning compared to full fine-tuning for adapting Stage \Romannum{1} to a new resolution. We then assess the effectiveness of RoPE in Stage \Romannum{2}. Next, we detail the low-quality video simulation strategy employed for training the Stage \Romannum{2} model. Additionally, we explore the importance of aligning the model's output with human preferences. Finally, we analyze the influence of various inference hyperparameters on the final performance.

\subsection{LoRA v.s. Full Parameter Fine-Tuning in Stage \Romannum{1}}

In the setup with a batch size of 32, we compare LoRA fine-tuning with full parameter fine-tuning for training the first-stage model at 270p resolution over the same number of iterations. The frame and video quality are evaluated on Texture100, and the semantics-related scores are assessed on VBench-Long, as shown in Table~\ref{table:lora}. In this configuration, full parameter fine-tuning tends to produce more artifacts, resulting in a degradation of both visual quality and semantic fidelity. In contrast, LoRA fine-tuning preserves the generative capabilities of the original model while efficiently adapting it to a lower resolution. Based on efficiency and performance, we opt for the LoRA strategy.

\begin{table*}[t]
\centering
\scriptsize
\setlength{\tabcolsep}{6pt}
\renewcommand{\arraystretch}{1.0}
\begin{tabular}{lccccccc}
\toprule
& \multicolumn{2}{c}{\textbf{Frame Quality}} & \multicolumn{2}{c}{\textbf{Video  Quality}} & \multicolumn{2}{c}{\textbf{Sematics}} \\
\cmidrule(lr){2-3}
\cmidrule(lr){4-5}
\cmidrule(lr){6-7}
 & MUSIQ($\uparrow$) & CLIPIQA($\uparrow$) & Technical($\uparrow$)   & Aesthetic($\uparrow$)  & Object Class($\uparrow$) & Overall Consistency($\uparrow$)   \\
\midrule
 Full Fine-Tuning & 20.53 & 0.273 & 8.531 & 97.64 & 85.6 & 26.1 \\
LoRA & \textbf{23.93} & \textbf{0.286} & \textbf{8.569} & \textbf{97.87} & 
\textbf{90.3} & \textbf{27.9}  \\
\bottomrule
\end{tabular}
\caption{Comparison of LoRA and full parameter fine-tuning in Stage \Romannum{1}. Best results  are in \textbf{bold}.}
\label{table:lora}
\end{table*}

\subsection{Position Embedding in Stage \Romannum{2}}
\label{sec:pos_emb}

\begin{figure}[t]
\begin{center}
\includegraphics[width=1\linewidth]{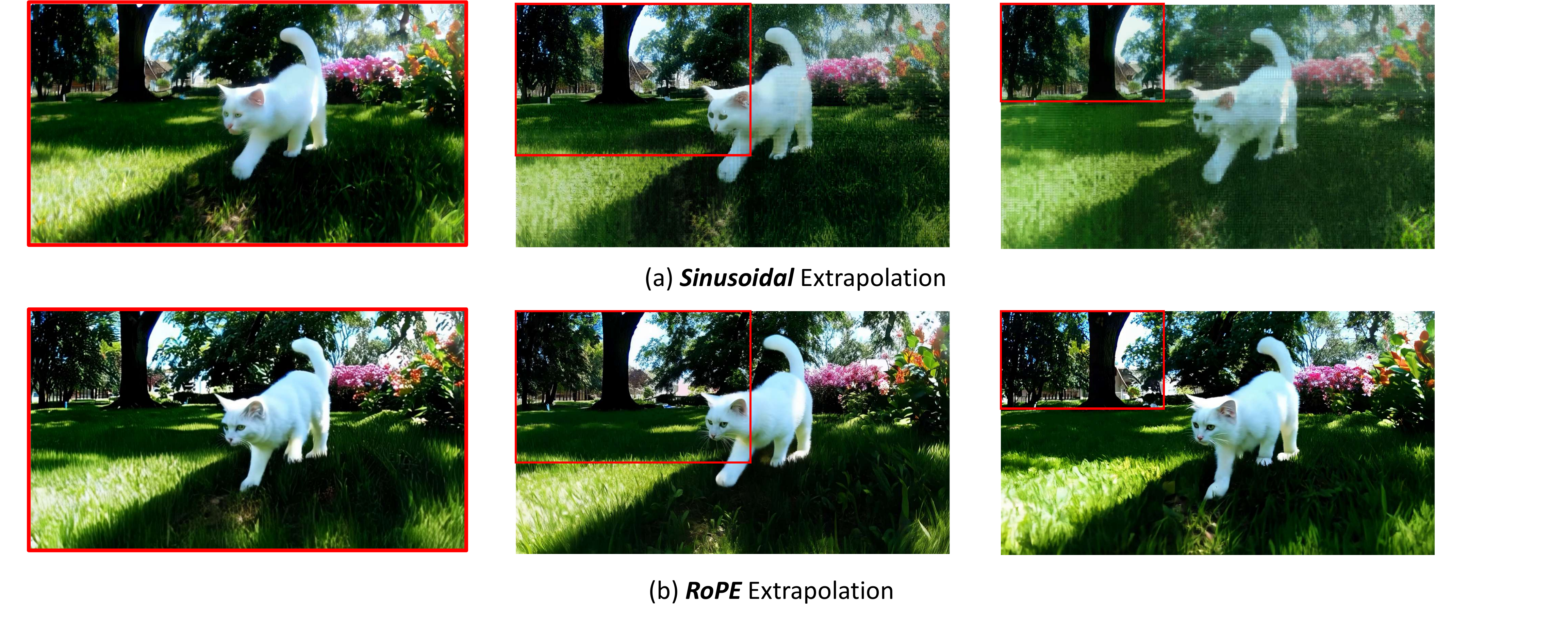}
\end{center}
\caption{\textbf{Results of resolution extrapolation using absolute sinusoidal and RoPE position embeddings.} Both settings perform well at the training resolution. However, while RoPE preserves detail enhancement at higher resolutions, absolute position embedding introduces noticeable artifacts beyond the training range.
}
\label{fig:rope}
\end{figure}

\begin{table*}[t]
\centering
\scriptsize
\setlength{\tabcolsep}{5pt}
\renewcommand{\arraystretch}{1.0}
\begin{tabular}{lccccccccc}
\toprule
&  & \multicolumn{4}{c}{\textbf{Frame Quality}} & \multicolumn{2}{c}{\textbf{Video  Quality}}  \\
\cmidrule(lr){3-6}
\cmidrule(lr){7-8}

    & \#NFE / Time &  MUSIQ($\uparrow$) & MANIQA($\uparrow$) & CLIPIQA($\uparrow$) & NIQE($\downarrow$)   &  Technical($\uparrow$)   & Aesthetic($\uparrow$)    \\
\midrule

FlashVideo-1080p & 4 / 72.2s &  58.69 & 0.296 & 0.439 & 4.501 & 11.86 & 98.92   \\
FlashVideo-2K  &  4 / 209.8s   & \textbf{62.40}  & \textbf{0.354} & \textbf{0.497} & \textbf{4.463} & \textbf{12.25} & \textbf{99.20}  \\

\bottomrule
\end{tabular}
\caption{Inference resolution scaling results of FlashVideo with RoPE. Best results  are in \textbf{bold}.}
\label{table:rope_extr}
\end{table*}

To achieve high training efficiency, we first train the Stage \Romannum{2} model at low resolution and then apply fine-tuning at higher resolutions, as detailed in Sec.\ref{sec:training}. Additionally, we aim for our model to generate high-quality videos at resolutions that exceed those used during training. To enable effective resolution generalization, we explore the use of representative position embeddings. Specifically, we compare the default absolute position embeddings~\citep{attention} from the 2-billion DiT model~\citep{cogvideox} with the rotary position embedding (RoPE)~\citep{rope}, and find that RoPE offers superior performance in such a video enhancement task.

We train the model using both position embeddings at a $540 \times 960$ resolution and test it across three settings: $540 \times 960$, $1080 \times 1920$, and $1440 \times 2560$. For the larger resolutions, we employ position embedding extrapolation. As shown in Figure~\ref{fig:rope}, while both position embeddings yield satisfactory results at the training resolution, RoPE consistently enhances details when inferring at larger scales. In contrast, absolute position embeddings exhibit clear artifacts beyond the trainining resolution. Based on these findings, we incorporate RoPE for training the second-stage model.

After training the model with RoPE at the 1080p ($1080 \times 1920$) resolution, we further extend the inference resolution to 2K ($1440 \times 2560$) using RoPE-based extrapolation. As shown in Table~\ref{table:rope_extr}, our model demonstrates improved visual quality at 2K resolution, as observed from the visual comparisons. However, the inference time increases significantly, from 74.4 seconds to 209.8 seconds. We hypothesize that larger resolutions better stimulate the detail-generation capabilities of our model, aligning with the inference scaling law~\citep{snell2024scaling} observed in large language models.

\subsection{Low-Quality Video Simulation in Stage \Romannum{2}}

\begin{table*}[!t]
\centering
\scriptsize
\setlength{\tabcolsep}{6pt}
\renewcommand{\arraystretch}{1.0}
\begin{tabular}{lccccccccc}
\toprule
  \multicolumn{2}{c}{\textbf{Degradation}} & \multicolumn{4}{c}{\textbf{Frame Quality}} & \multicolumn{2}{c}{\textbf{Video  Quality}}  \\
\cmidrule(lr){1-2} \cmidrule(lr){3-6} \cmidrule(lr){7-8}
 $DEG_{pixel}$ & $DEG_{latent}$ & MUSIQ($\uparrow$) & MANIQA($\uparrow$) & CLIPIQA($\uparrow$) & NIQE($\downarrow$) & Technical($\uparrow$) & Aesthetic($\uparrow$) & \\
\midrule
  &  & 23.61   & 0.200   & 0.286 & 12.02  & 6.43 & 97.32 \\
 \checkmark &  & 49.12   & 0.253 & 0.364 & 4.95  & 7.12 & 99.02\\

 \checkmark & \checkmark & \textbf{55.45} & \textbf{0.273} & \textbf{0.409} & \textbf{4.69} & \textbf{9.09} & \textbf{98.96}  \\

\bottomrule
\end{tabular}
\caption{Comparison of frame quality and video quality when applying different degradations. Best results are in \textbf{bold}.}
\label{table:deg}
\end{table*}

\label{sec:deg_discuss} As discussed in Section~\ref{exp:stage2}, we visually demonstrate (see Figure~\ref{fig:deg_ab}) the significance of incorporating latent and pixel degradation for simulating low-quality videos during the training of Stage \Romannum{2}. In this section, we provide a more detailed quantitative evaluation. For computational efficiency, we conduct the experiment using 5-frame 1080p video inputs. We train two models for 10,000 iterations: one with only pixel degradation applied, and the other with both pixel and latent degradation. As shown in Table~\ref{table:deg}, the baseline represents the results from Stage \Romannum{1}. When the Stage \Romannum{2} model is applied with pixel degradation ($DEG_{pixel}$), the first-stage output is significantly improved, with high-frequency textures being added and overall visual quality boosted. Furthermore, incorporating latent degradation ($DEG_{latent}$) leads to even further enhancement, producing clearer and more realistic structures for small objects and background details.

\subsection{Human Preference Alignment in Stage \Romannum{2}}
\begin{table*}[!t]
\centering
\scriptsize
\setlength{\tabcolsep}{5pt}
\renewcommand{\arraystretch}{1.0}
\begin{tabular}{lcccccc}
\toprule
  & \multicolumn{4}{c}{\textbf{Frame Quality}} & \multicolumn{2}{c}{\textbf{Video  Quality}}  \\
\cmidrule(lr){2-5}
\cmidrule(lr){6-7}
   &  MUSIQ($\uparrow$) & MANIQA($\uparrow$) & CLIPIQA($\uparrow$) & NIQE($\downarrow$)  & Technical($\uparrow$) & Aesthetic($\uparrow$)  \\
\midrule
Before   & 55.61  & 0.278 & 0.427 & 4.667 & 11.76 & 98.90   \\
After   &  \textbf{58.69} & \textbf{0.296} & \textbf{0.439} & \textbf{4.501} & \textbf{11.86} & \textbf{98.92}  \\

\bottomrule
\end{tabular}
\caption{Performance comparison of FlashVideo before and after human preference alignment.  Best results  are in \textbf{bold}.}\label{table:human_stage}
\end{table*}

In our experiments, training at 1080p resolution reveals instability, characterized by performance fluctuations across different checkpoints (every 500 iterations). We attribute this inconsistency to the varying quality of the training samples. To address this issue, we manually curate a high-quality dataset of 50,000 samples, specifically selected based on strong human preference. Our model undergoes a quick fine-tuning process on this refined dataset to stabilize training and improve performance, and then is evaluated on the Texture100 benchmark, as presented in Table~\ref{table:human_stage}. Despite the relatively small size of the selected dataset, we observe substantial improvements in both aesthetic quality and the richness of fine details. These results highlight the effectiveness of incorporating human preference into the fine-tuning process.

% \subsection{Inference Hyperparameters}

\begin{table*}[!t]
\tiny
\setlength{\tabcolsep}{0.6pt}
\centering
\begin{minipage}{0.45\textwidth}
\centering
\renewcommand{\arraystretch}{1.0}

\scalebox{1}{
\begin{tabular}{lccccccc}
\toprule
& \multicolumn{4}{c}{\textbf{Frame Quality}} & \multicolumn{2}{c}{\textbf{Video  Quality}}  \\
\cmidrule(lr){2-5}\cmidrule(lr){6-7}
  NFE &  MUSIQ($\uparrow$) & MANIQA($\uparrow$) & CLIPIQA($\uparrow$) & NIQE($\downarrow$)  & Tech($\uparrow$) & Aesth($\uparrow$) \\
\midrule

1 & 48.60 & 0.253 & 0.307 & 5.148 & 8.643 & 98.03  \\
2 & 55.10 & 0.287 & 0.390 & 4.730 & 10.57 & 98.38  \\
3 & 57.59 & 0.290 & 0.418 & 4.543  & 11.39 &  98.62  \\
\rowcolor{gray!40} 4 &  58.69 & 0.296 & 0.439 & 4.501 & 11.86 & 98.92  \\
\rowcolor{gray!40} 5 & 59.24 & 0.299 & 0.441 & 4.492 & 12.15 & 99.05  \\
\rowcolor{gray!40} 6 & 59.17 & 0.295 & 0.440 & 4.521 & 12.48 & 99.05 \\
7 & 59.48 & 0.298 & 0.445 & 4.578 & 12.20 & 99.01  \\
8 & 59.64 & 0.298 & 0.451 & 4.554 & 12.05  & 99.16  \\

\bottomrule
\end{tabular}
}
\caption{Results of FlashVideo under different numbers of function evaluations (NFEs). The recommended range is highlighted in gray.}\label{table:nfe}
\end{minipage}%
\hspace{0.4cm}
\begin{minipage}{0.45\textwidth}
\centering
\renewcommand{\arraystretch}{1.0}
\scalebox{1}{
\begin{tabular}{lcccccccc}
\toprule
& \multicolumn{4}{c}{\textbf{Frame Quality}} & \multicolumn{2}{c}{\textbf{Video  Quality}}  \\
\cmidrule(lr){2-5}\cmidrule(lr){6-7}
CFG &  MUSIQ($\uparrow$) & MANIQA($\uparrow$) & CLIPIQA($\uparrow$) & NIQE($\downarrow$)  & Tech($\uparrow$) & Aesth($\uparrow$) \\

\midrule

1 & 45.01 & 0.253 & 0.359 & 5.395 & 10.75& 98.98 \\
4 & 50.92 & 0.278 & 0.397 & 5.102 &  11.71 & 99.16  \\
7 & 54.26 & 0.287 & 0.418 & 4.905 & 11.97 & 99.10   \\
 \rowcolor{gray!40}{10} & 57.37 & 0.298 & 0.441 & 4.692 & 12.15 & 99.12  \\ 
\rowcolor{gray!40}{13} &  58.69 & 0.296 & 0.439 & 4.501 & 11.86 & 98.92  \\
16 & 58.42 & 0.285 & 0.416 & 4.353 & 11.54 & 98.57  \\
19 & 57.66 & 0.277 & 0.397 & 4.143 & 11.32 & 97.84 \\
22 & 57.48 & 0.270 & 0.379 & 3.982 & 10.94  & 97.76  \\

\bottomrule
\end{tabular}
}
\caption{Results of FlashVideo under different classifier-free guidance (CFG) scales. The recommended range is highlighted in gray.  }\label{table:cfg}
\end{minipage}

\end{table*}
\begin{table*}[!t]
\scriptsize
\setlength{\tabcolsep}{6pt}
\centering
\begin{tabular}{lccccccc}
\toprule
 & & \multicolumn{4}{c}{\textbf{Frame Quality}} & \multicolumn{2}{c}{\textbf{Video  Quality}}  \\
\cmidrule(lr){3-6}\cmidrule(lr){7-8}
Training Noise Step & Inf Noise &  MUSIQ($\uparrow$) & MANIQA($\uparrow$) & CLIPIQA($\uparrow$) & NIQE($\downarrow$)  & Tech($\uparrow$) & Aesth($\uparrow$) \\
\midrule
 \multirow{6}{*}{600-900}& 600 & 53.62 & 0.269 & 0.403 & 4.911 & 11.85 & 99.03 \\
 &  650 & 53.98 & 0.269 & 0.399 & 4.832 & 11.77 & 99.06  \\
 &  700  & 53.82 & 0.274 & 0.399 & 4.763 & 11.93 & 99.02 \\
 &  750 & 54.06 & 0.279 & 0.400 & 4.785 & 11.96 & 98.92  \\
& 800 & 53.50 & 0.276 & 0.403 & 4.663 & 11.72 & 98.91 \\
& 850 & 51.39 & 0.279 & 0.391 & 4.787 & 11.26 & 98.72  \\
\midrule
 \multirow{5}{*}{650-750}& 650 & 58.49 & 0.294 & 0.431 & 4.583 & 11.96 & 98.84  \\
& 675  & 58.69 & 0.296 & 0.439 & 4.501 & 11.86 & 98.92\\
&700 & 57.80 & 0.290 & 0.418 & 4.531 & 12.01 & 98.78  \\
&725 & 57.97 & 0.295 & 0.426 & 4.462 & 11.98 & 98.83 \\
&750 & 57.62 & 0.294 & 0.422 & 4.437 & 12.10 & 98.72 \\
\bottomrule
\end{tabular}
\centering
\caption{Results of FlashVideo under different latent degradation strengths. During initial training, a noise step range of 600–900 is applied, with model performance evaluated across different steps. The range of 650–750 consistently yields satisfactory results (see upper half of Table). This refined range is then adopted for subsequent training, with final performance presented in the lower half of Table.}\label{table:latent_deg}
\end{table*}

During the testing phase, users can flexibly adjust several hyperparameters—namely the number of function evaluations (NFEs), classifier-free guidance (CFG), and latent degradation strength (noise strength)—to suit their specific needs. We provide a detailed analysis of how these hyperparameters affect performance in Figure~\ref{fig:ab_inf}, with corresponding quality scores reported in Tables~\ref{table:nfe},~\ref{table:cfg}, and~\ref{table:latent_deg}. Unless otherwise specified, the default values for these hyperparameters are set to NFE=4, CFG=13, and NOISE=675.

\subsection{Number of Function Evaluations} As depicted in Figure~\ref{fig:ab_inf}~(a), the processed video exhibits slight haziness and blurriness when NFE=1. Increasing the NFE improves visual quality, with more defined facial details, \textit{e.g.}, teeth and hair, and sharper textures on elements such as leaves and sweaters observed at NFE=4. Beyond NFE=4, increasing the value further (\textit{i.e.}, to NFE=5 or higher) does not result in significant visual enhancement in most cases. The qualitative results on some metrics reported in Table~\ref{table:nfe} confirm this trend, aligning with the visual observations. We recommend users to adjust the NFE to between 4 and 6 during actual use.

\subsection{Classifier-free Guidance} The impact of the CFG scale is illustrated in Figure~\ref{fig:ab_inf}~(b). At CFG=1, the result remains blurry, with insufficient details. As the CFG value increases, the video content becomes clearer and more defined, with finer details such as earrings becoming more distinctly visible. Specifically, CFG values between 10 and 13 yield satisfactory results, striking a balance between sharpness and details. However, further  increasing CFG beyond 13 results in excessive sharpness, leading to unnaturally textured visuals. As shown in Table~\ref{table:cfg}, both image and video quality scores improve as CFG increases from 1 to 13, but several metric scores degrade when CFG exceeds 13.

\subsection{Latent Degradation Strength} 
\begin{figure}[!t]
\begin{center}
\includegraphics[width=1\linewidth]{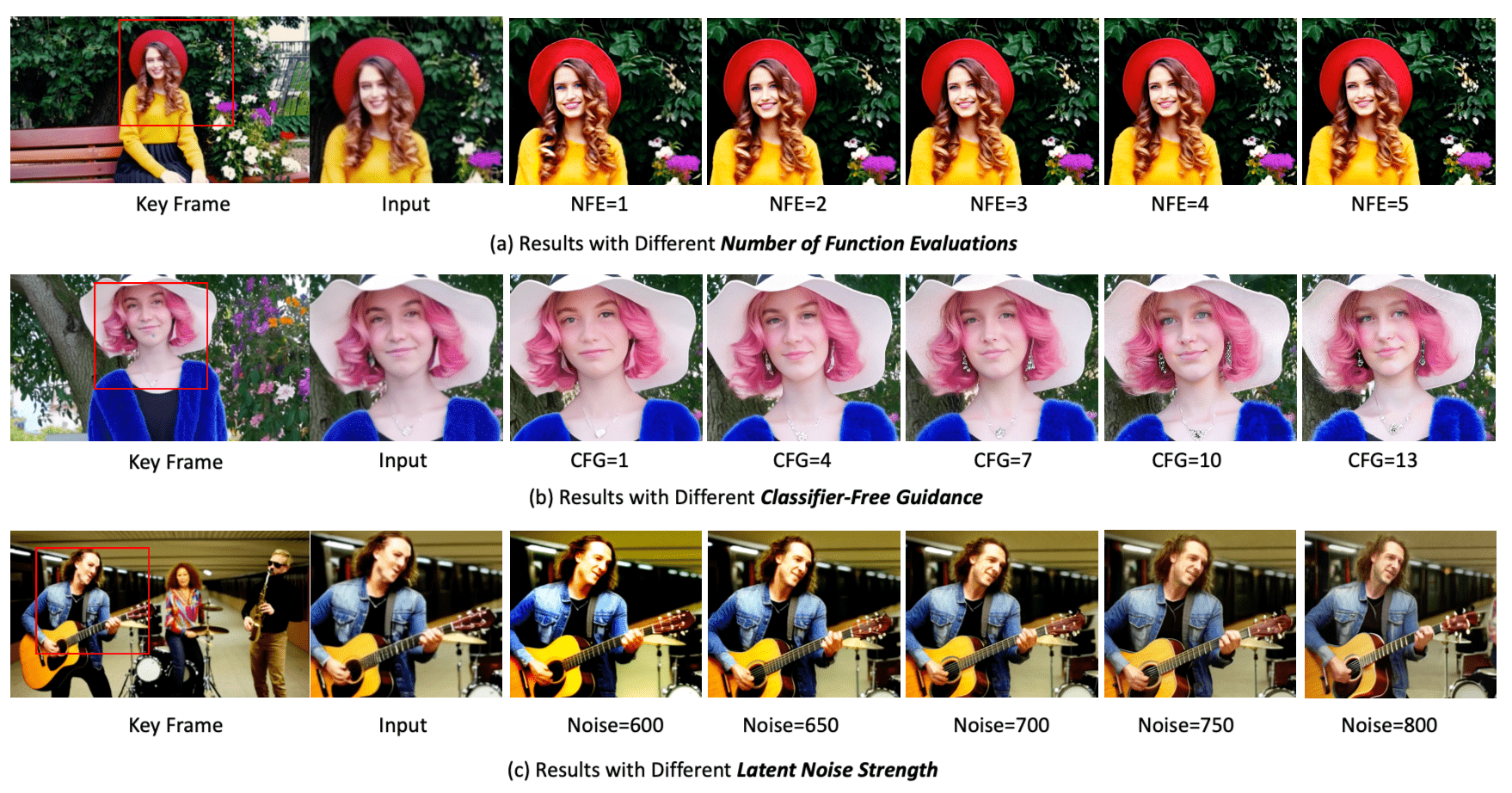}
\end{center}
\caption{Results of stage \Romannum{2} under different inference hyper-parameters. }
\label{fig:ab_inf}
\end{figure}

We find that latent degradations below 500 noise steps lead the model to ``cheat'' and fail to correct first‑stage artifacts. To identify an effective strength, we conduct a noise‑step search over the 600–900 interval (upper half of Table~\ref{table:latent_deg}) and observe that the 650–750 window consistently delivers the best artifact removal and visual content preservation. Accordingly, we adopt the 650–750 noise‑step range for our final training regime (results shown in the lower half of Table~\ref{table:latent_deg}). The qualitative visualizations further support this analysis. As shown in Figure~\ref{fig:ab_inf}~(c), at lower degradation levels, the enhanced video retains higher fidelity to the original input. This preservation of fidelity, while beneficial for maintaining overall content integrity, can impede the repair of artifacts and restrict the generation of finer details, such as those seen in fingers, guitar strings, and surface textures. On the other hand, increasing the noise strength promotes the generation of additional visual details. Yet, if the noise is excessive, it can distort structures or introduce blurriness, due to the inherent limitations of Stage \Romannum{2}'s generative capacity.

\section{Discussion and Limitation}

\subsection{Discussion} 

In this section, we share some insights from our exploration to help readers gain a clearer understanding of the design principles and positioning of our work, as well as to provide guidance for potential future improvements.

\paragraph{Principles of adjusting latent degradation strength.} Selecting an appropriate latent degradation strength is crucial for training the Stage \Romannum{2} model. Achieving the balance between minimizing artifacts and preserving the integrity of the original content is key. We recommend adjusting the latent degradation strength based on the Signal-to-Noise Ratio (SNR), meaning that the noise step should be increased when either the resolution or the number of video frames increases. Notably, the number of frames has a greater impact than resolution, as visual content across multiple frames exhibits stronger correlations that are harder to disrupt. For example, in preliminary experiments with 17 video frames, we find that artifacts in the input could be corrected with a noise step of 500, which is significantly lower than the optimal noise range of 650 to 750 observed when the frame count is increased to 49.

\paragraph{Fidelity vs. visual quality improvement.} A delicate balance exists between maintaining fidelity and enhancing visual quality. Unlike real-world video enhancement, where input videos purely lack high-frequency details, the first-stage generated video often contains subtle structural flaws or artifacts that require refinement. Traditional super-resolution methods, which focus on maintaining high fidelity, are unable to address these issues effectively. Conversely, regenerating new content by treating the first-stage output as a rough guide also falls short, as it conflicts with our design philosophy. We view the first-stage output as a low-cost preview, and it must align closely with the final result. To achieve this balance, we carefully adjust the strength of both strategies, ensuring that visual quality is enhanced without compromising the integrity of the original content.

\paragraph{Can Stage \Romannum{2} be a general video enhancement model?} It is noteworthy that the current training setup is specifically tailored for 1080p  and is not suitable as a general enhancement method for videos with varying resolutions or frame counts. However, we believe that with further refinement, such as incorporating additional input information regarding resolution and frame number, the model could be adapted to handle a wider range of scenarios. We aim to explore this direction in future work.
\paragraph{Challenges with increased video length.} Video enhancement is more challenging than single-image processing, as it requires ensuring the consistency of newly added details across the entire video sequence. This task calls for a model that not only improves visual quality but also manages the intricate visual relationships and motion across frames. In Stage \Romannum{2}, we address these challenges by employing 3D full attention and adjusting the degradation strength. However, as the video length increases, the computational demand of 3D full attention escalates quadratically. Moreover, if the degradation strength is not carefully adjusted, the model may resort to recovering details by directly referencing multiple frames, which can compromise its generative capacity during inference.
\paragraph{Sparse attention in Stage \Romannum{2}.} We visualize the attention maps in Stage \Romannum{2} and observe significant sparsity, particularly in space compared to time. We attribute this phenomenon to the moderate motion intensity in the current first-stage output. To reduce the computational cost of Stage \Romannum{2}, we apply FlexAttention~\citep{flexattention} to implement window-based spatial-temporal attention with $H=11, W=11, T=7$. As a result, the method performs well with significantly improved efficiency when the first-stage output contains low motion. However, we observe inconsistencies and blurred patterns in the regenerated visual details when motion is large. We propose that dynamically adjusting the window size based on motion intensity could be a promising solution in future work.

\paragraph{Resolutions of two stages.} Given sufficient computational resources, higher resolutions in both stages could be pursued. Our choice of 270p for the first stage is driven by its ability to produce preliminary results in only 30 seconds, allowing users to quickly assess whether further computation in Stage \Romannum{2} is necessary. This provides a clear advantage over contemporary methods.

\subsection{Limitation}

\paragraph{Time-Consuming VAE decoding for high-resolution videos.} Due to GPU memory constraints, decoding 1080p videos requires spatial and temporal slicing, a process that is time-consuming. Engineering advances in parallel processing and more efficient VAE architectures are essential for enabling faster generation of high-resolution videos.

\paragraph{Long Prompt for inference.} The text descriptions adopted during training are typically long and highly detailed. This may increase complexity when users provide prompts in inference. Future research could employ joint training with short prompts or engage language models designed for prompt rewriting~\citep{ji2024prompt}. This advancement can significantly enhance the user experience.

\paragraph{Challenges with fast motion.} Due to constraints in data quantity, quality, and diversity, Stage \Romannum{2} may fail when processing videos with extreme and fast motion. Potential solutions include incorporating more training data with large motion and scaling up the model capacity.

\section{Conclusions}

We introduce FlashVideo, a novel two-stage framework that separately optimizes prompt fidelity and visual quality. This decoupling allows for strategic allocation of both model capacity and the number of function evaluations (NFEs) across two resolutions, greatly enhancing computational efficiency. In the first stage, FlashVideo prioritizes fidelity at a low resolution, utilizing large parameters and sufficient NFEs. The second stage performs flow matching between low and high resolutions, efficiently generating fine details with fewer NFEs. Extensive experiments and ablation studies demonstrate the effectiveness of our approach. Moreover, FlashVideo delivers preliminary results at a very low cost, enabling users to decide whether to proceed to the enhancement stage. This decision-making capability can significantly reduce costs for both users and service providers, offering substantial commercial value.

\clearpage
{
    \small
    \bibliography{reference}
    \bibliographystyle{flashvideog}

}

\end{document}